\documentclass[10pt,twocolumn,letterpaper]{article}

\usepackage[pagenumbers]{cvpr} 

\usepackage{graphicx}
\usepackage{amsmath}
\usepackage{amssymb}
\usepackage{booktabs}
\usepackage{multirow}
\usepackage{rotating}
\usepackage{array}
\usepackage{adjustbox}
\usepackage{caption}
\usepackage[flushleft]{threeparttable}

\usepackage[pagebackref,breaklinks,colorlinks]{hyperref}

\usepackage[capitalize]{cleveref}
\crefname{section}{Sec.}{Secs.}
\Crefname{section}{Section}{Sections}
\Crefname{table}{Table}{Tables}
\crefname{table}{Tab.}{Tabs.}

\def\cvprPaperID{*****} 
\def\confName{CVPR}
\def\confYear{2023}
\newcommand{\taskName}{\texttt{Multimodal Egocentric Scene Modeling}\xspace}
\newcommand{\algoName}{\texttt{Aria-NeRF}\xspace}
\newcommand{\datasetName}{\texttt{Aria-NeRF Dataset}\xspace} 

\newcolumntype{L}[1]{>{\raggedright\let\newline\\\arraybackslash\hspace{0pt}}m{#1}}
\newcolumntype{C}[1]{>{\centering\let\newline\\\arraybackslash\hspace{0pt}}m{#1}}
\newcolumntype{R}[1]{>{\raggedleft\let\newline\\\arraybackslash\hspace{0pt}}m{#1}}
\begin{document}

\title{\algoName: Multimodal Egocentric View Synthesis}

\author{Jiankai Sun$^\dag$, Jianing Qiu$^\ddag$, Chuanyang Zheng$^\ddag$, John Tucker$^\dag$, Javier Yu$^\dag$,  Mac Schwager$^\dag$\\
$^\dag$Stanford University  $^\ddag$The Chinese University of Hong Kong}

\maketitle

\begin{abstract}
We seek to accelerate research in developing rich, multimodal scene models trained from egocentric data, based on differentiable volumetric ray-tracing inspired by Neural Radiance Fields (NeRFs).  The construction of a NeRF-like model from an egocentric image sequence plays a pivotal role in understanding human behavior and holds diverse applications within the realms of VR/AR.  Such egocentric NeRF-like models may be used as realistic simulations, contributing significantly to the advancement of intelligent agents capable of executing tasks in the real-world. The future of egocentric view synthesis may lead to novel environment representations going beyond today's NeRFs by augmenting visual data with multimodal sensors such as IMU for ego-motion tracking, audio sensors to capture the surface texture and human language context, and eye-gaze trackers to infer human attention patterns in the scene.  To support and facilitate the development and evaluation of egocentric multimodal scene modeling, we present a comprehensive multimodal egocentric video dataset. This dataset offers a comprehensive collection of sensory data, featuring RGB images, eye-tracking camera footage, audio recordings from a microphone, atmospheric pressure readings from a barometer, positional coordinates from GPS, connectivity details from Wi-Fi and Bluetooth, and information from dual-frequency IMU datasets (1kHz and 800Hz) paired with a magnetometer. The dataset was collected with the Meta Aria Glasses wearable device platform. We evaluated two baseline NeRF-based models, Nerfacto and NeuralDiff, on our dataset. While they were capable of producing reasonable visual reproduction of the scene, our findings also highlight opportunities for further improvement using a variety of sensing modalities beyond vision. The diverse data modalities and the real-world context captured within this dataset serve as a robust foundation for furthering our understanding of human behavior and enabling more immersive and intelligent experiences in the realms of VR, AR, and robotics.
\end{abstract}

\section{Introduction}
\label{sec:intro}
\begin{figure*}
    \centering
    \includegraphics[width=\linewidth]{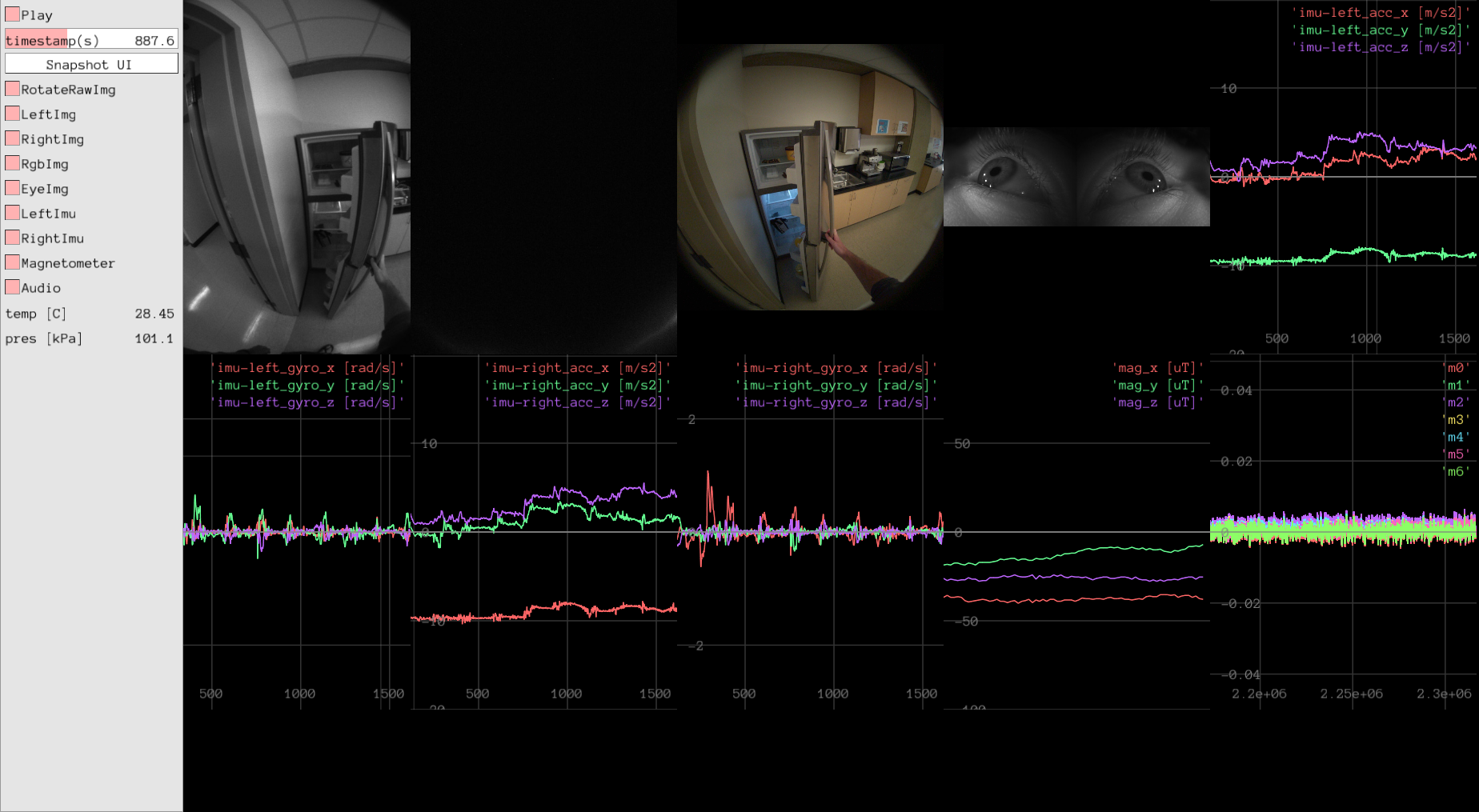}
    \caption{\datasetName, Kitchen 1 subset, comprises a diverse range of sensory data, including RGB images, ET camera, microphone, barometer, GPS, Wi-Fi, Bluetooth, SLAM, and two sets of IMU data (1kHz and 800Hz), along with a magnetometer.}
    \label{fig:dataset_intro}
\end{figure*}

Recent advances in VR/AR technologies highlight a growing need for creating immersive virtual environments. Neural Radiance Fields (NeRFs)~\cite{mildenhall2020nerf} is a technique that has gained much attention for its capability to generate photorealistic 3D scenes, meeting this demand for strengthened immersion and realism.
Utilizing NeRF for the creation of lifelike simulations is of significant value, particularly in the development of intelligent agents capable of executing real-world tasks. 
Nevertheless, dynamic NeRF remains a complex problem~\cite{li2023dynibar}. Unlike traditional NeRF, which deals with static scenes, dynamic NeRF aims to capture and represent objects and scenes that change over time. This introduces the need for modeling complex temporal dynamics, such as object motion, deformation, or interactions, which can be challenging to represent accurately. On the other hand, it also presents an intriguing avenue to explore whether multimodal sensory data can enhance NeRF training. This work focuses on egocentric view synthesis, which is a natural scenario featuring rich multimodal data that can be captured by multi-sensory wearable devices.

The progression of egocentric vision~\cite{qiu2022egocentric,ramakrishnan2023naq} heavily relies on hardware advancements. This is particularly relevant in the context of wearable devices like Aria Glasses~\cite{pan2023aria, somasundaram2023project}, which capture data from a first-person perspective in real-life scenarios.  Aria Glasses, designed as a research tool to accelerate advances in AR/VR, embodied AI, and human behavior modeling, employ a range of sensors to capture first-person perspective video, audio, data on eye movement and location, providing a comprehensive platform for understanding a user's intention, as well as their interactions with the world. 

In this paper, we present the \datasetName\footnote{Dataset: \url{https://tinyurl.com/aria-nerf}}, a comprehensive multimodal egocentric dataset designed for multimodal egocentric scene modeling, with diverse real-world multi-sensory data captured using Aria Glasses. Extensive experiments demonstrate that our dataset is a rich testbed for typical NeRF-based tasks and algorithms, with high-quality annotations to explore by future research. Our dataset is also scalable, offering a cost-effective pipeline for converting Aria Glasses-captured videos into NeRF-compatible training data.

In summary, our main contributions are as follows:
\begin{itemize}
    \item We introduce the task of \taskName, a step towards egocentric view synthesis with neural scene representations, using Fisheye RGB images and multimodal sensory data.
    \item We build a novel \datasetName, which includes multiple modalities, such as Fisheye images, RGB, depth, IMU, audio, etc. The proposed \datasetName serves as a rich testbed for advancing multimodal NeRF, for example, audio-guided NeRF, and gaze-guided NeRF. In particular, \datasetName has language annotations, which are suitable for training LLMs-guided NeRF.
    \item  We evaluate and benchmark Nerfacto~\cite{nerfstudio} and NeuralDiff~\cite{tschernezki2021neuraldiff} on the proposed \datasetName extensively. The results reveal the challenging nature of our dataset and egocentric view synthesis, suggesting the need for further improvement of the current NeRF methods.
\end{itemize}
\begin{table*}[t]
\begin{center}
\renewcommand\tabcolsep{0.5pt}
\begin{tabular}{>{\centering\arraybackslash}m{0.17\linewidth}>{\centering\arraybackslash}m{0.17\linewidth}>{\centering\arraybackslash}m{0.17\linewidth}>{\centering\arraybackslash}m{0.17\linewidth}>{\centering\arraybackslash}m{0.17\linewidth}}
\toprule
Lounge-1 & Lounge-2 &  Lounge-3  & Conference-Room-1 & Conference-Room-2   \\
\midrule
\includegraphics[width=\linewidth]{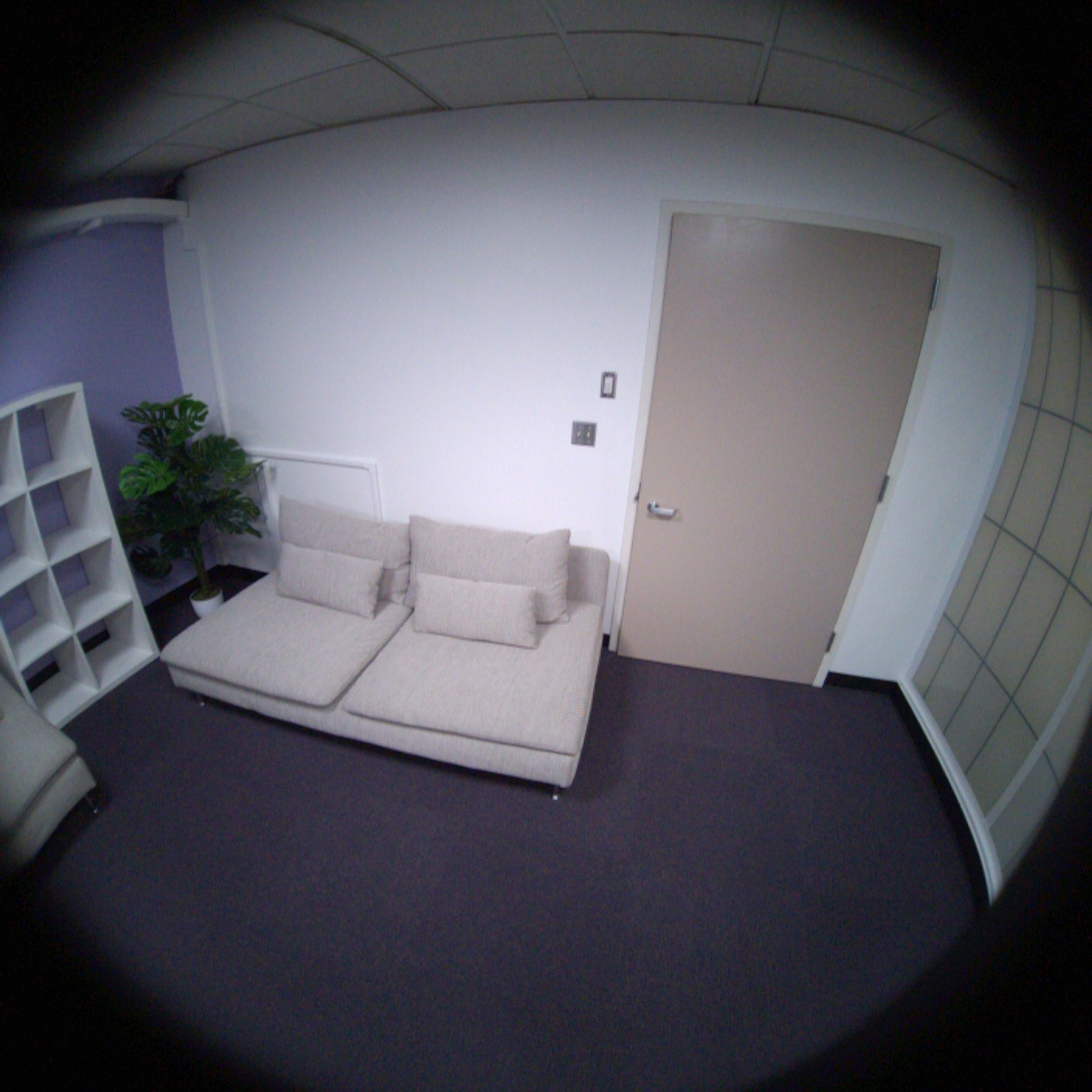} & 
\includegraphics[width=\linewidth]{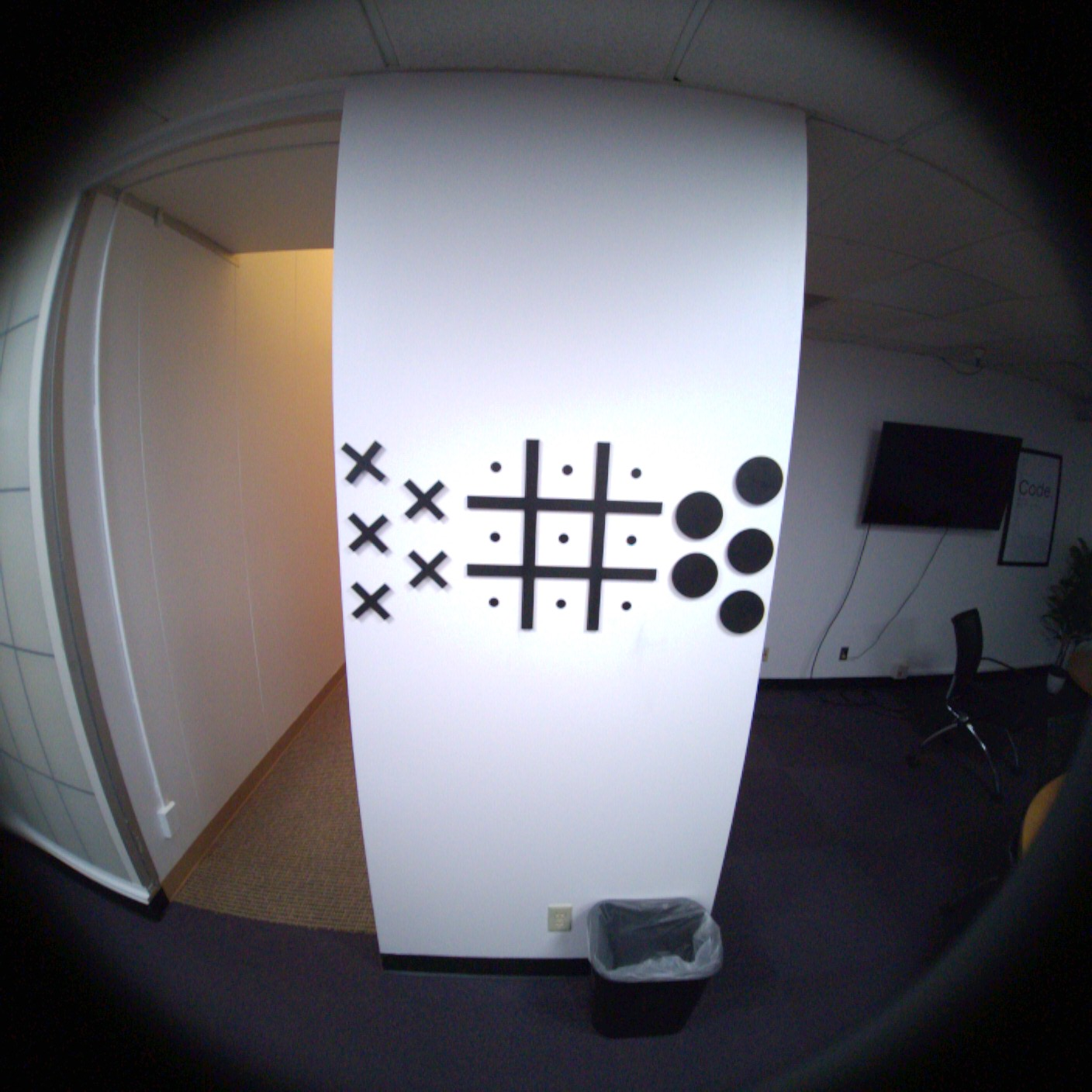} & 
\includegraphics[width=\linewidth]{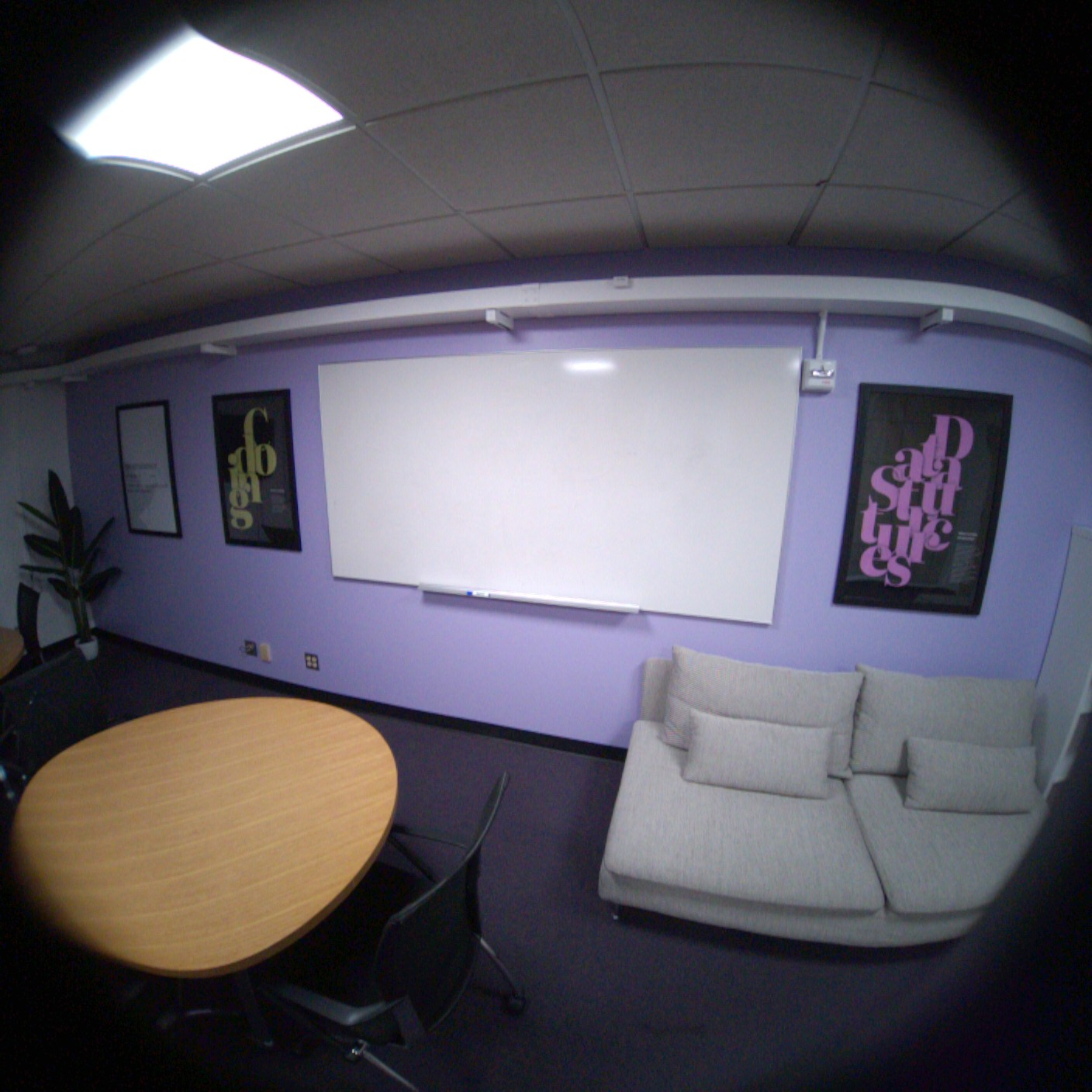} & 
\includegraphics[width=\linewidth]{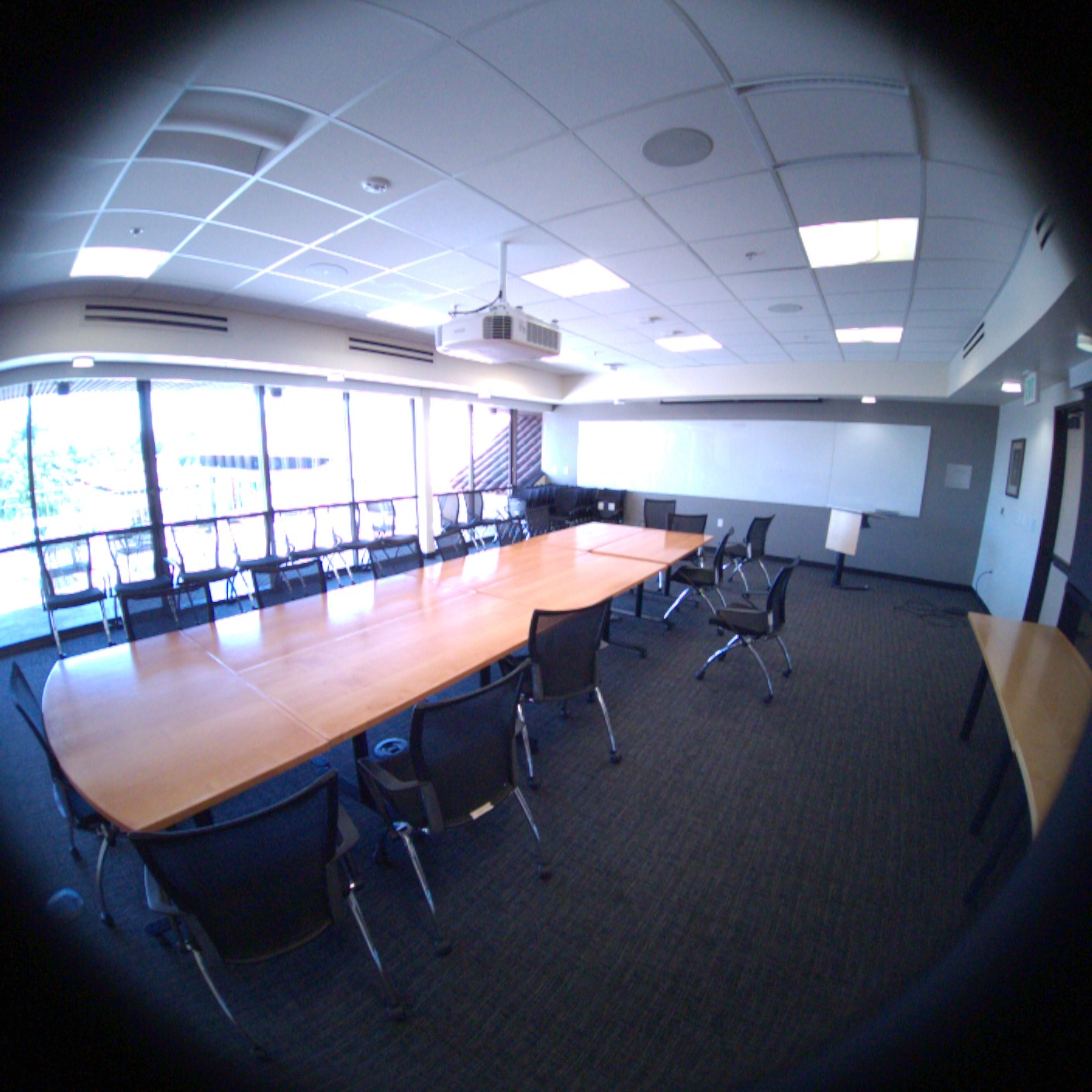} & \includegraphics[width=\linewidth]{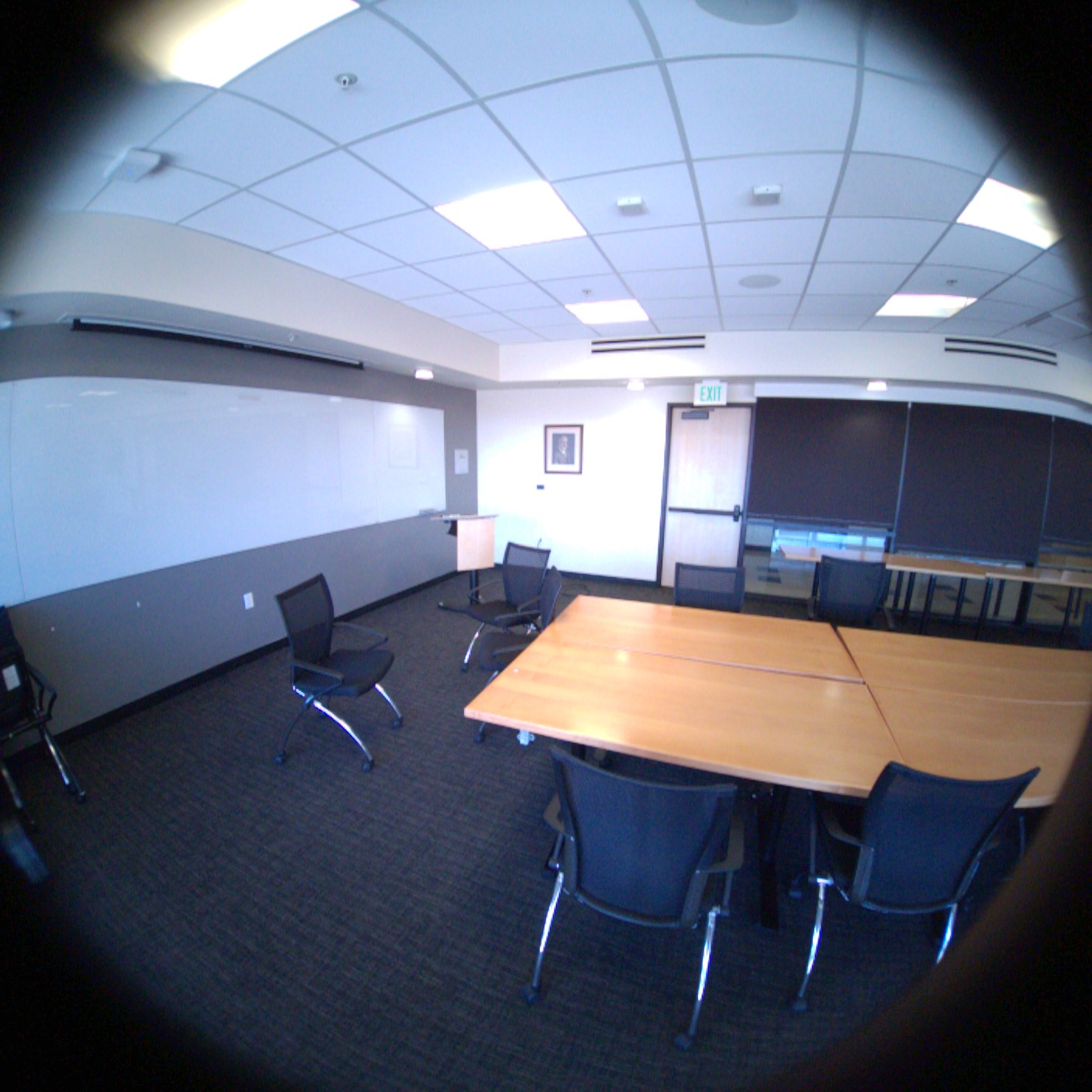} \\ 
\midrule
Drone-Workbench-1 & Drone-Workbench-2 & Drone-Workbench-3 & Kitchen-1  & Kitchen-2  \\
\midrule
\includegraphics[width=\linewidth]{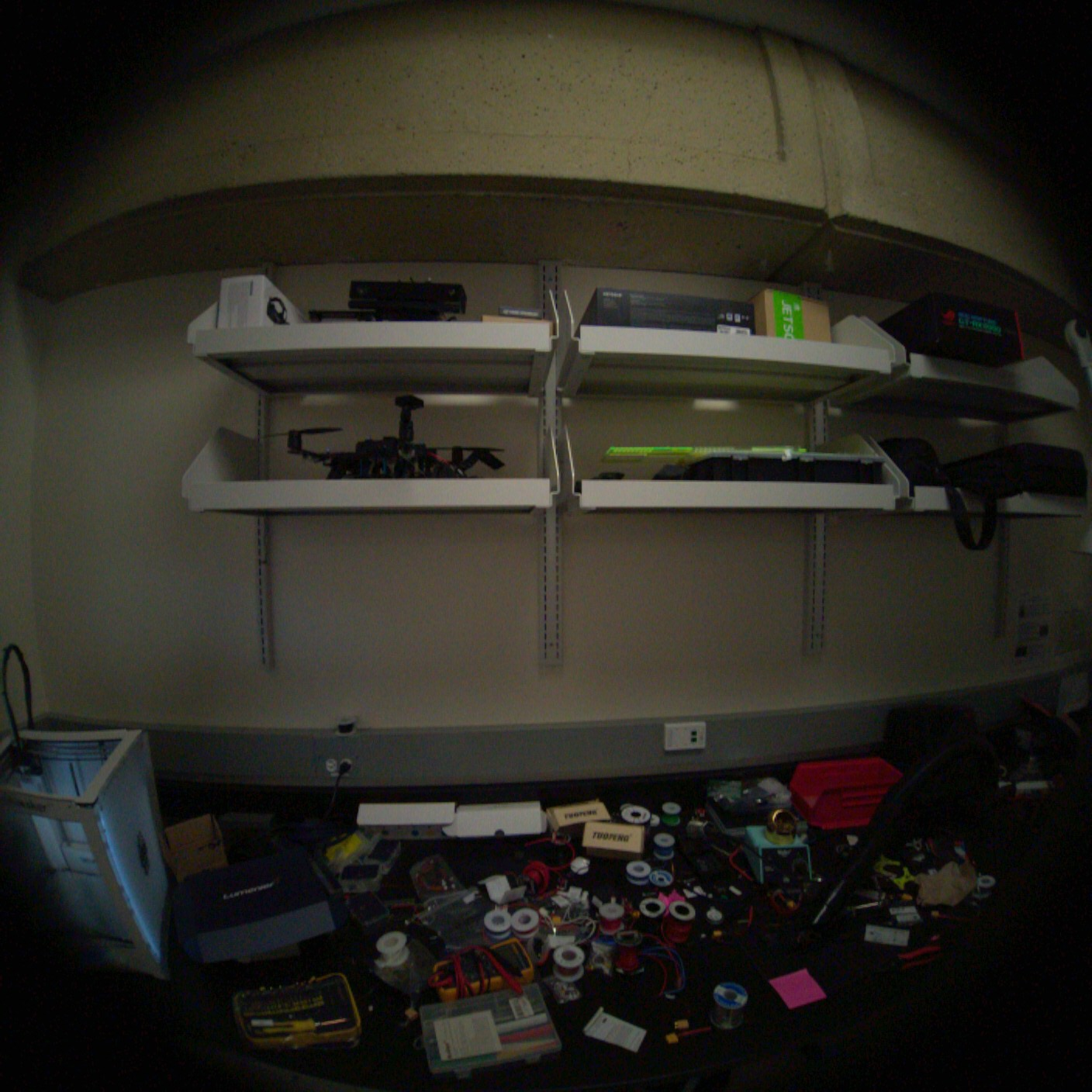} &
\includegraphics[width=\linewidth]{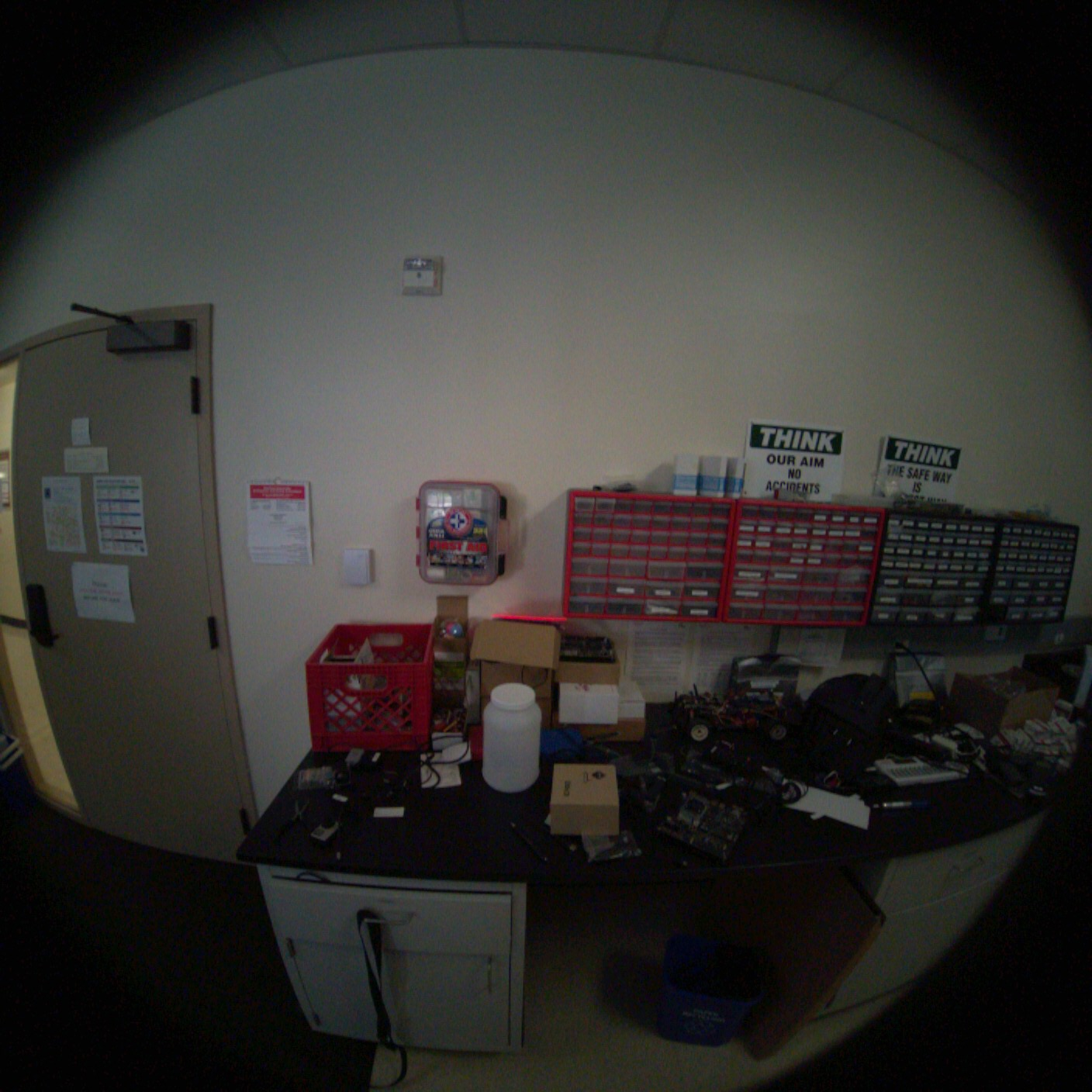}   & 
\includegraphics[width=\linewidth]{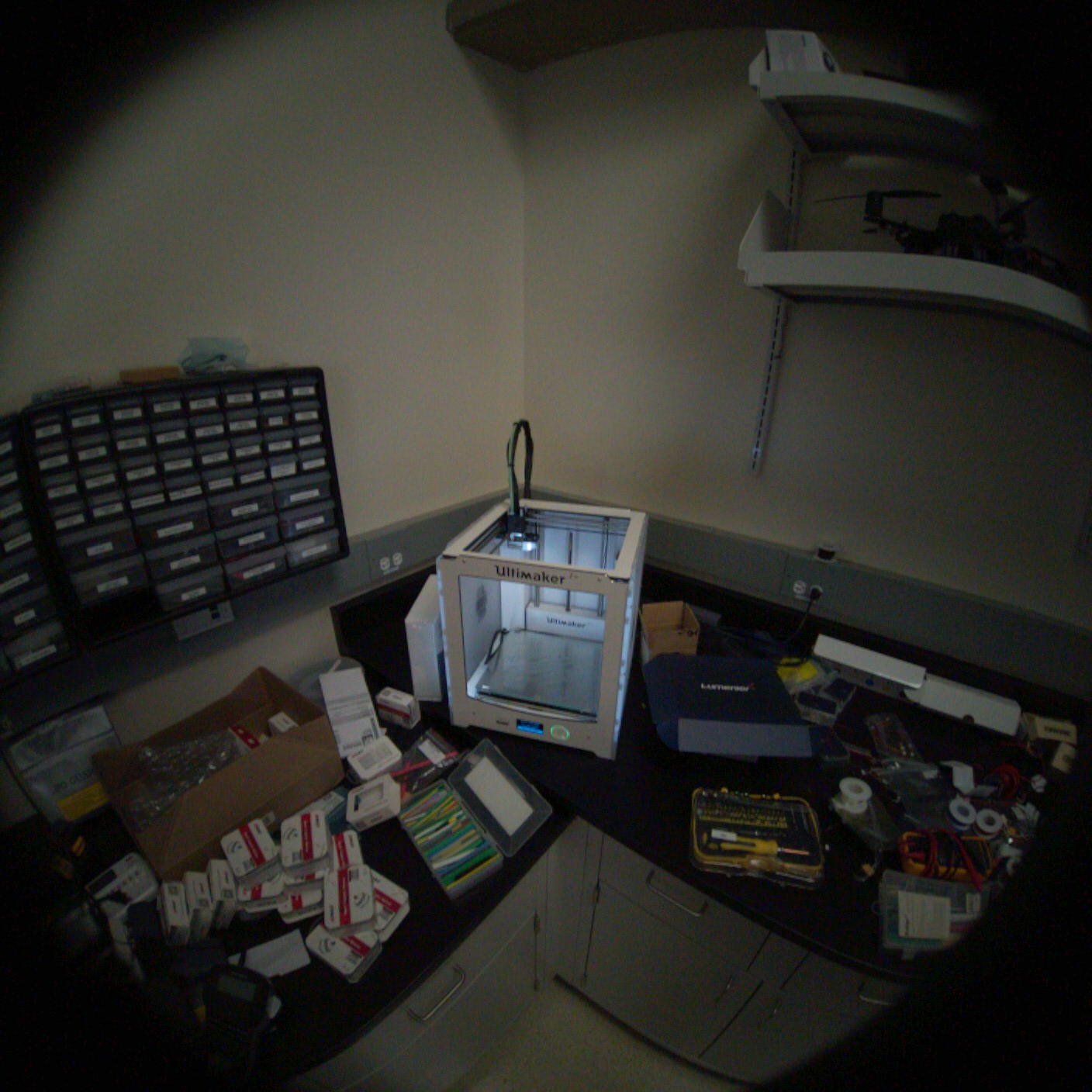}   & \includegraphics[width=\linewidth]{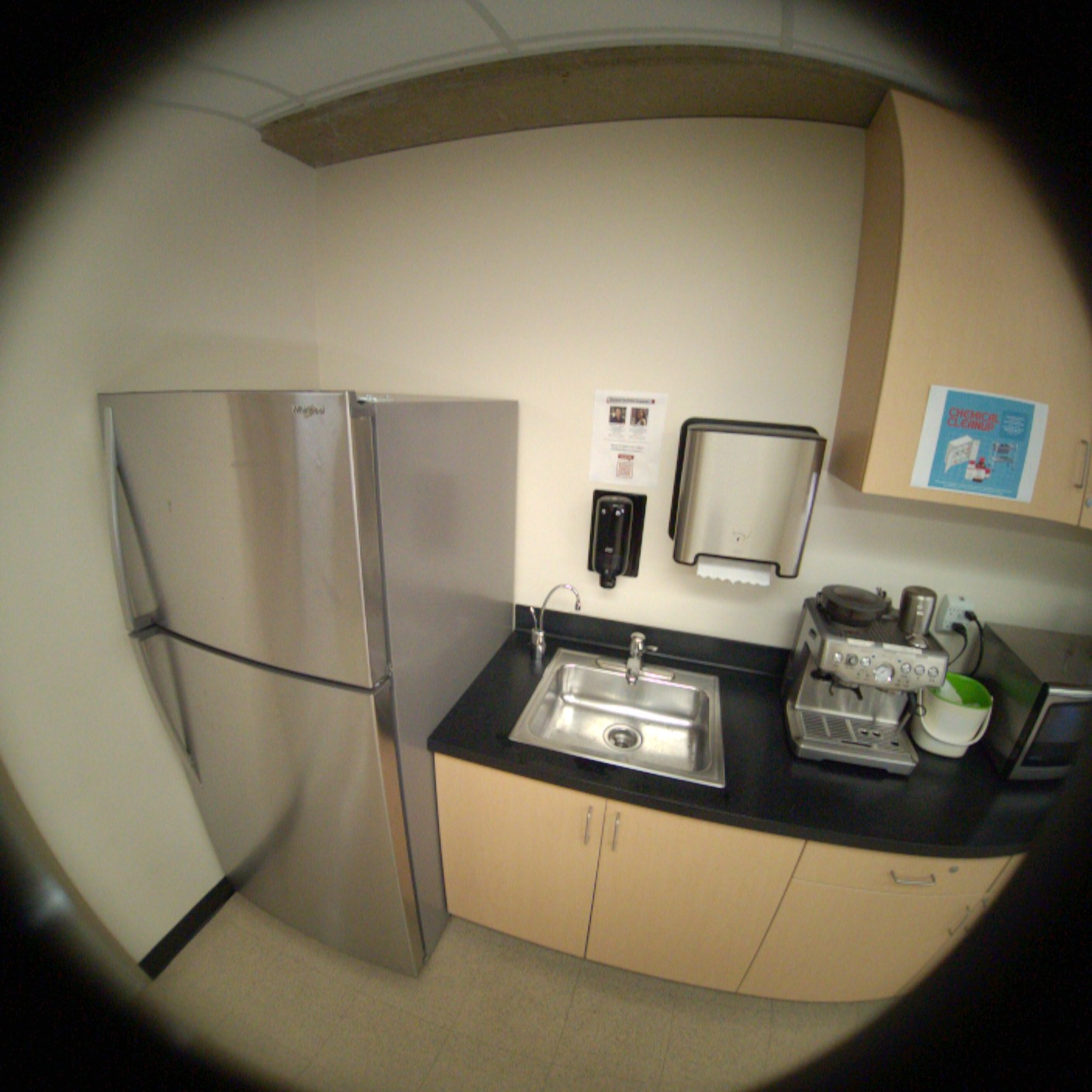} & \includegraphics[width=\linewidth]{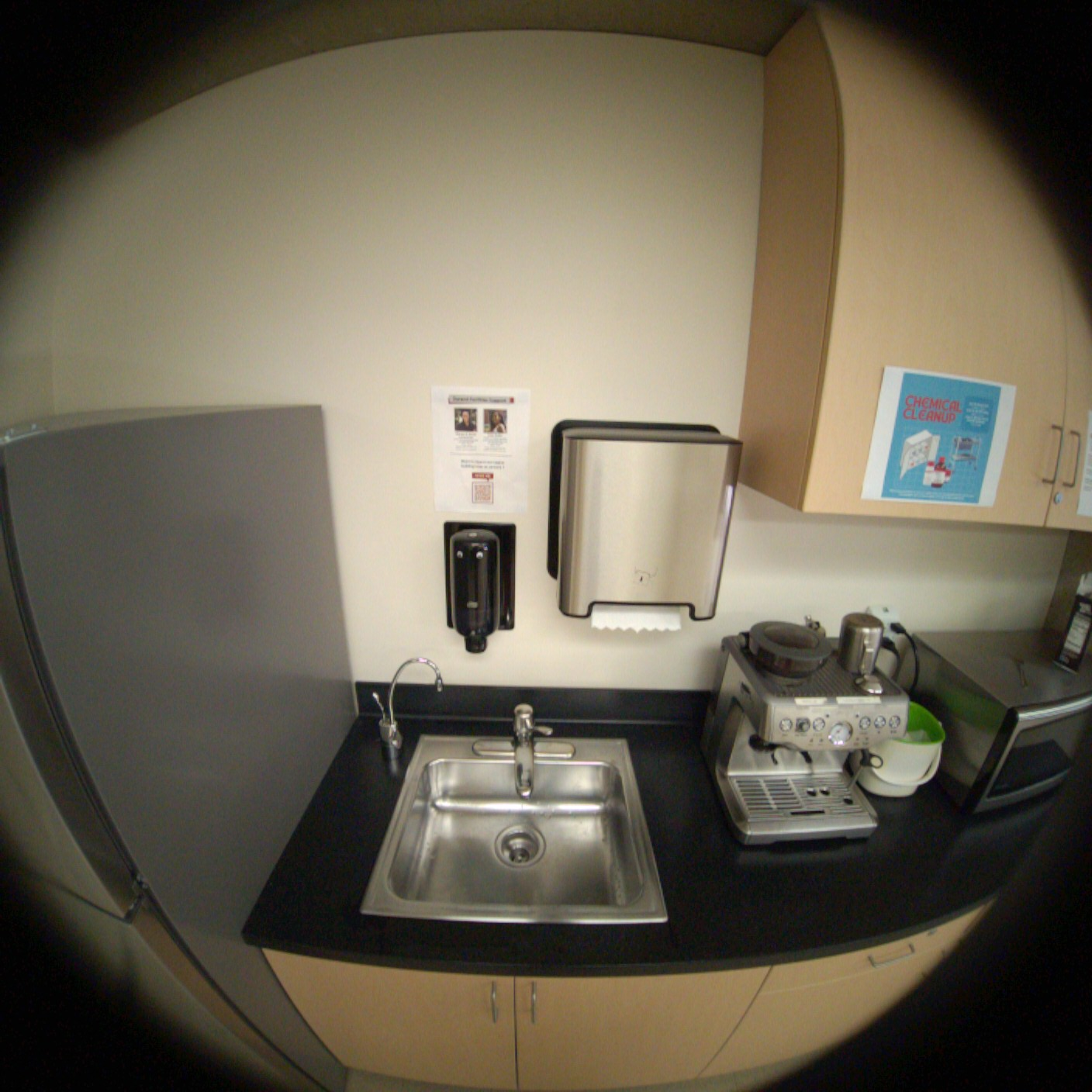} 
\\
\midrule
Robot-Workbench-1 &  Robot-Workbench-2 & MSL Flightroom & Fountain 1 \\
\midrule
\includegraphics[width=\linewidth]{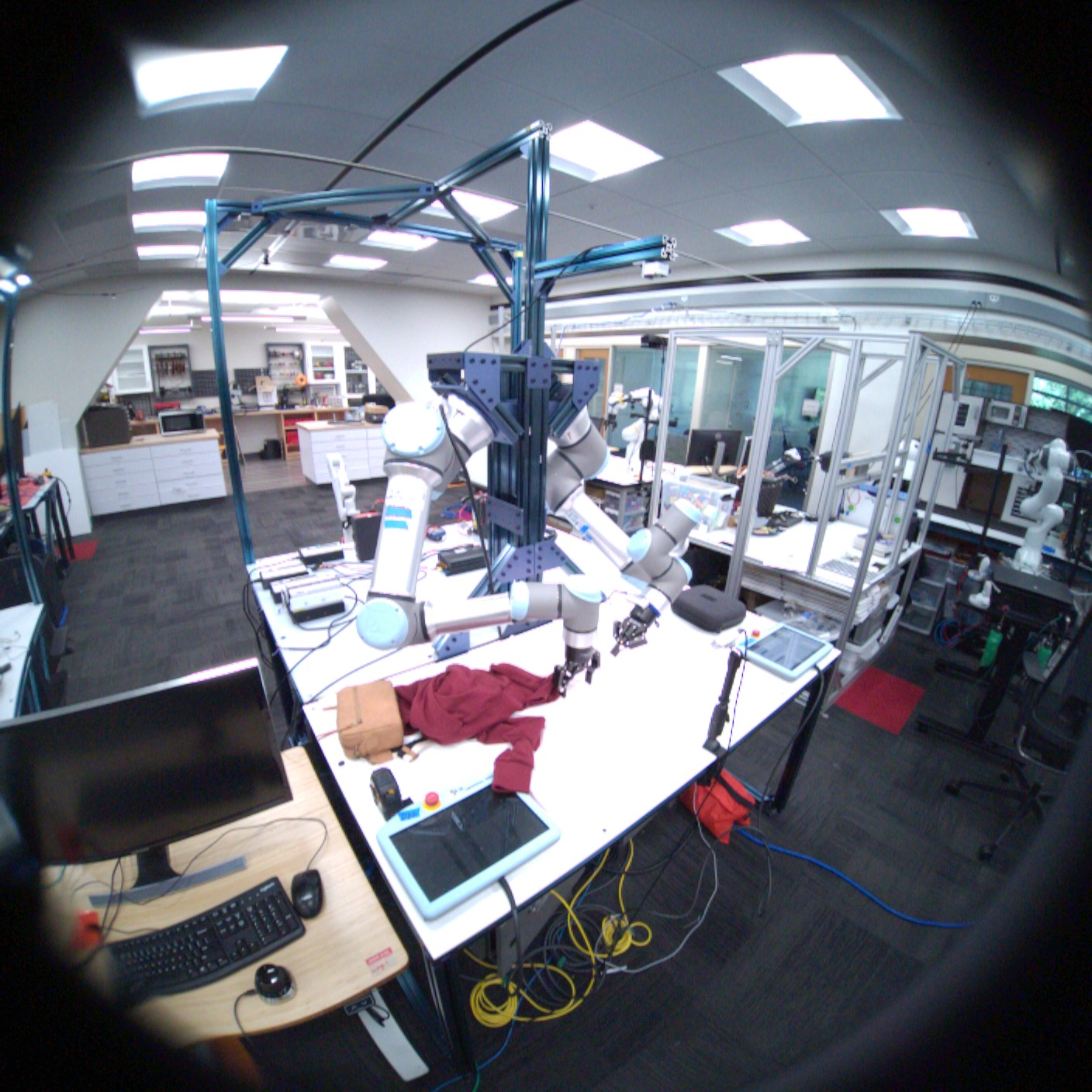} &
\includegraphics[width=\linewidth]{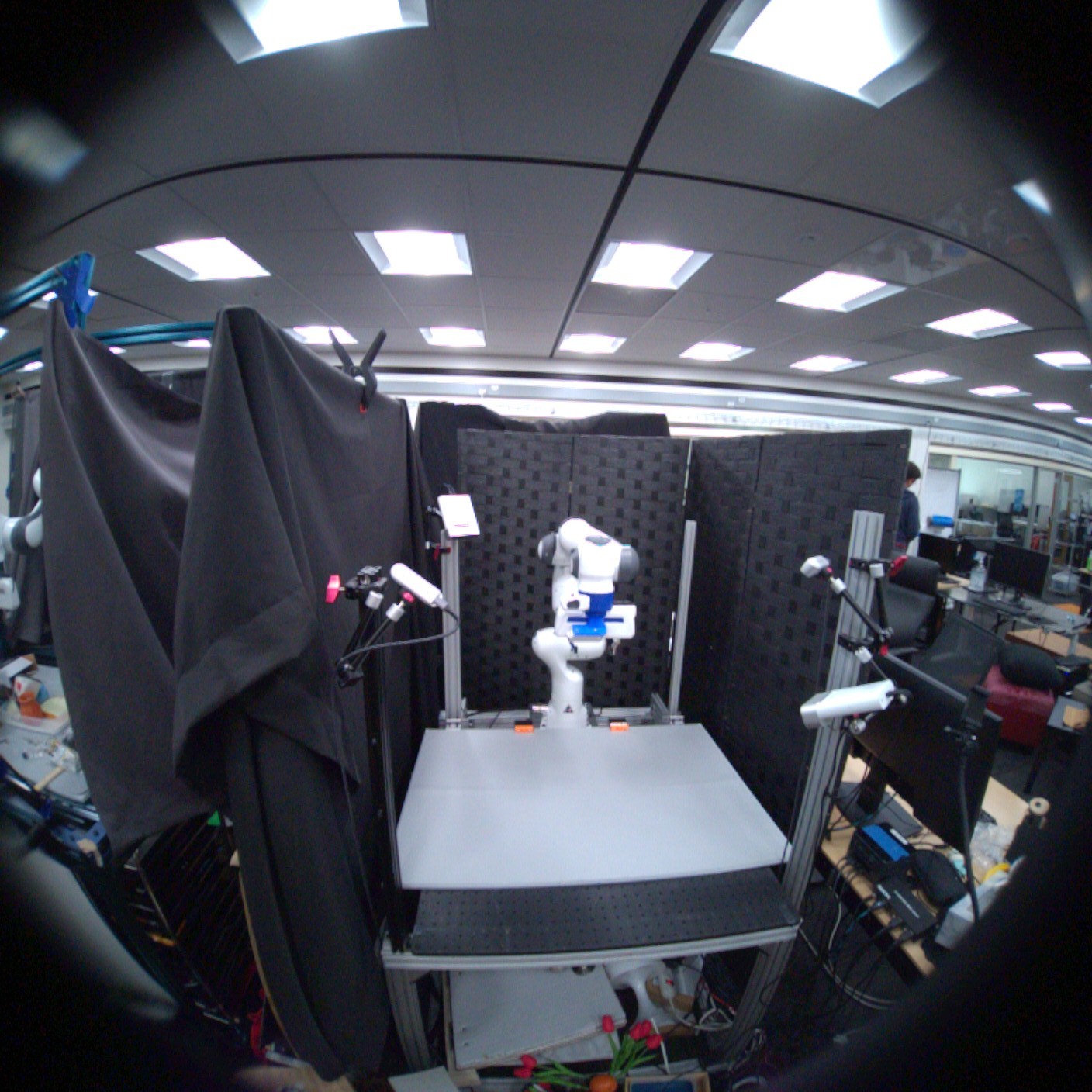} &
\includegraphics[width=\linewidth]{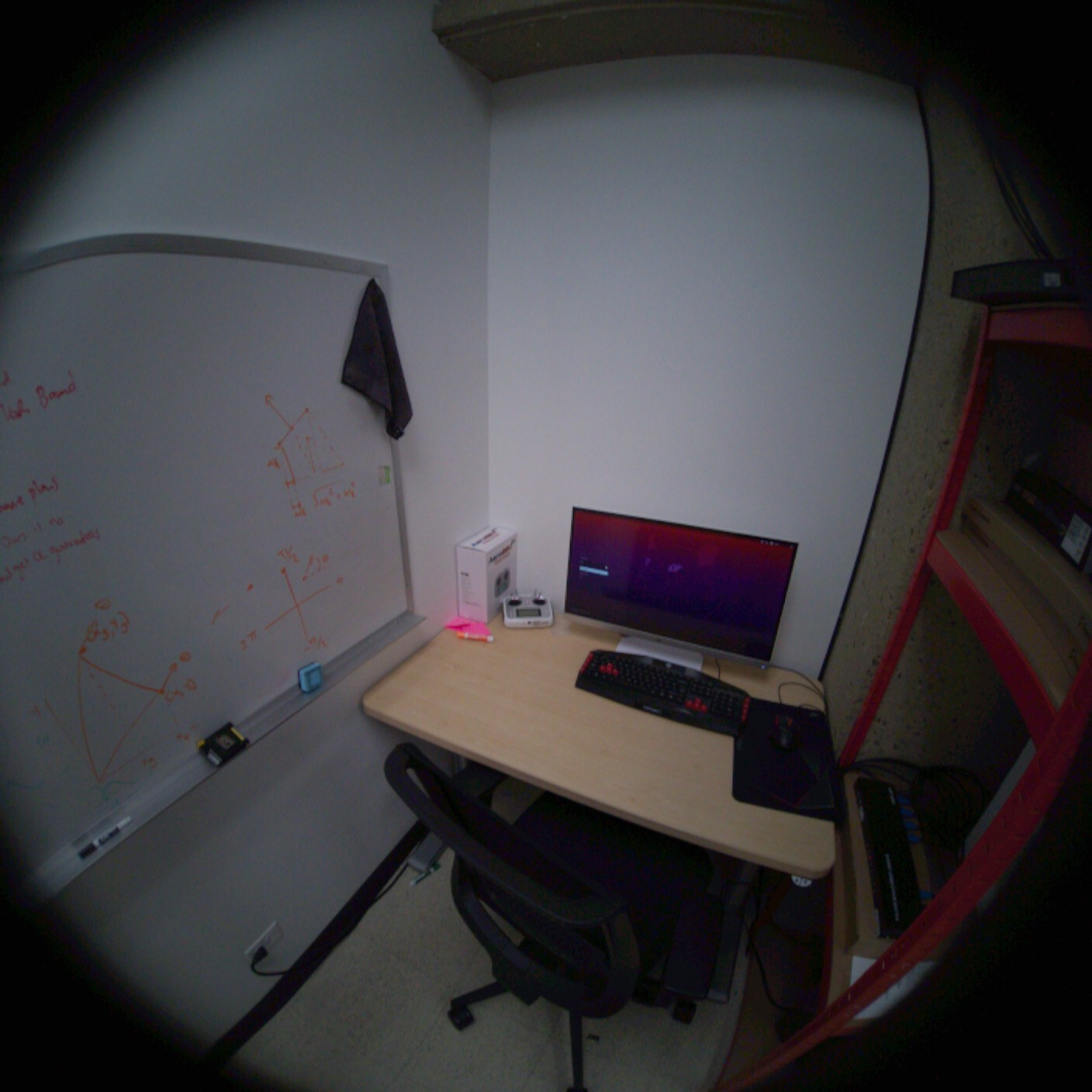} &
\includegraphics[width=\linewidth]{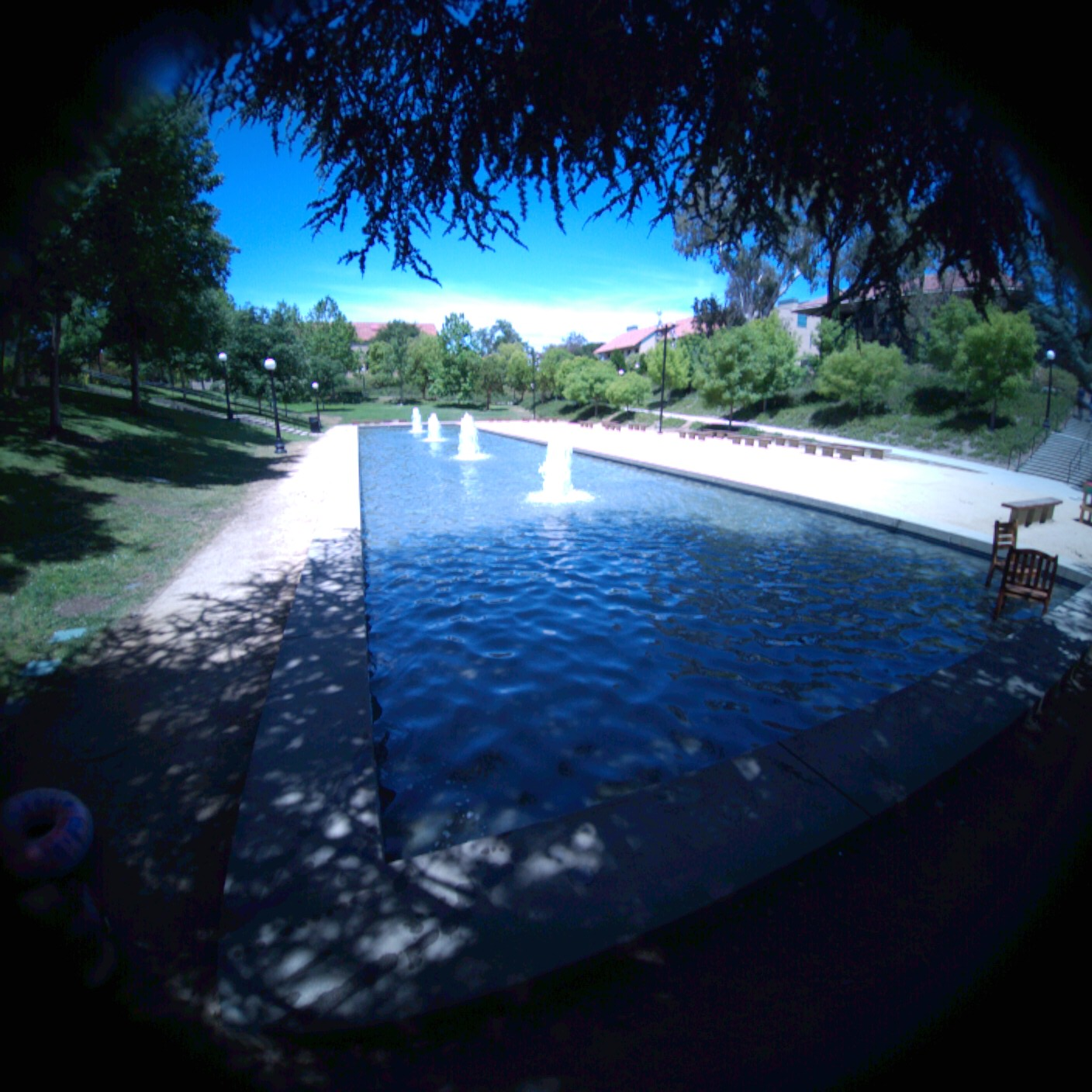} 
\\
\bottomrule
\end{tabular}
\captionof{figure}{Scene Examples}
\label{tab:all_scene_example}
\end{center}
\end{table*}

\begin{table*}[]
    \centering
    \caption{Dataset Statistics 1: Time and Sensors - RGB, Eye Tracking (ET), Microphone, Barometer, and GPS Data.}
    \label{tab:dataset_stat1}
    \begin{tabular}{ccccccc}
    \toprule
   Subset & Time (s) & RGB & ET & Microphone & Barometer & GPS  \\
       \midrule
       Fountain1  & 124.9 & 1261 & 1261 & 2953 & 6276 & 127  \\
       Lounge-1  & 69.0 & 701 & 701 & 1640 & 3497 & 0 \\
       Lounge-2  & 69.6 & 707 & 707 & 1656 & 3524 & 0  \\
       Lounge-3  & 146.5 & 1476 & 1476 & 3457 & 7374 & 0  \\
       Flightroom  & 159.6 & 1607 & 1607 & 3765 & 8046 & 0 \\
       Drone-Workbench-1  & 145.3 & 1465 & 1465 & 3430 & 7316 & 0  \\
       Drone-Workbench-2  & 127.5 & 1286 & 1287 & 3013 & 6411 & 0  \\
       Drone-Workbench-3  & 78.2 & 793 & 793 & 1856 & 3960 & 0   \\
       Conference-Room-1  & 107.4 & 1085 & 1085 & 2541 & 5412 & 0 \\
       Conference-Room-2  & 77.2 & 783 & 783 & 1833 & 3907 & 0  \\
       Kitchen-1  & 134.8 & 1360 & 1360 & 3185 & 6791 & 0  \\
       Kitchen-2  & 123.4 & 1245 & 1245 & 2917 & 6216 & 0  \\
       Robot-Workstation-1  & 69.7 & 708 & 708 & 1657 & 3530 & 71 \\
       Robot-Workstation-2  & 41.5  & 426 & 426 & 996 & 2122 & 43 \\     
       \bottomrule
    \end{tabular}
\end{table*}

\begin{table*}[]
    \centering
    \caption{Dataset Statistics 2: encompasses a diverse array of sensory data including Wi-Fi, Bluetooth, SLAM, and measurements from both IMUs and magnetometers. The dataset features two variants of IMU data: one sampled at 1kHz and another at 800Hz.}
    \label{tab:dataset_stat2}
    \begin{tabular}{ccccccc}
    \toprule
   Subset  & Wi-Fi & Bluetooth & SLAM & IMU (1kHz) & IMU (800Hz) & Magnetometer  \\
       \midrule
       Fountain1   & 340 & 1 & 1261 & 126029 & 102124 & 1262 \\
       Lounge-1  & 326 & 26 & 700 & 70037 & 56751 & 702 \\
       Lounge-2  & 294 & 37 & 707 & 70683 & 57275 & 708 \\
       Lounge-3  & 583 & 81  & 1476 & 147492 & 119508 & 1480 \\
       Flightroom  & 476 & 0 & 1607 & 160611 & 130143 & 1613 \\
       Drone-Workbench-1  & 530 & 0 & 1465 & 146314 & 118556 & 1468 \\
       Drone-Workbench-2  & 408 & 0 & 1286 & 128586 & 104195 & 1288  \\
       Drone-Workbench-3  & 250 & 0 & 793 & 79241 & 64211 & 794  \\
       Conference-Room-1  & 391 & 34 & 1085 & 108440 & 87871 & 1086 \\
       Conference-Room-2  & 157 & 20 & 783 & 78257 & 63410 & 784 \\
       Kitchen-1  & 522 & 0 & 1360 & 135900 & 110119 & 1363 \\
       Kitchen-2  & 606 & 0 & 1245 & 124453 & 100845 & 1248  \\
       Robot-Workstation-1  & 543 & 0 & 708 & 70760 & 57338 & 708 \\
       Robot-Workstation-2  & 318 & 0 & 426 & 42562 & 34490 & 426 \\     
       \bottomrule
    \end{tabular}
\end{table*}

\begin{table*}[]
    \centering
    \caption{Quantitative Results. In the context of PSNR, SSIM, and LPIPS metrics, NeuralDiff generally surpasses Nerfacto across various scenarios. }
    \label{tab:quan_results}
    \begin{threeparttable}
    \begin{tabular}{c|c|c|c|c|c|c}
    \toprule
   \multirow{2}{*}{Subset} & \multicolumn{3}{c|}{Nerfacto} & \multicolumn{3}{c}{NeuralDiff} \\
    \cline{2-7}
        & PSNR$\uparrow$ & SSIM $\uparrow$  &  LPIPS $\downarrow$  & PSNR$\uparrow$ & SSIM $\uparrow$  &  LPIPS $\downarrow$ \\
       \midrule
       Fountain1  & 20.16 & 0.7075 & 0.5212 &  29.09 & 0.9131 & 0.1695 \\
       Lounge-1 & 19.93 & 0.7284 & 0.5261 & 29.98 & 0.9270 & 0.2238  \\
       Lounge-3 & 19.63 & 0.7036 & 0.5951 & 29.93 & 0.9230 & 0.1441 \\
       Flightroom &  19.41 & 0.7177 & 0.4601 & 29.30 & 0.8979 & 0.1689 \\
       Drone-Workbench-1  & 20.52 & 0.6820 & 0.5626 & 25.67 & 0.8578 & 0.3315 \\
       Drone-Workbench-2 & 20.93  & 0.6887 & 0.5169 & 30.60 & 0.9324 & 0.1149 \\
       Drone-Workbench-3 & 25.05 & 0.7736 & 0.4804 & 31.28 & 0.9427 & 0.1050 \\
       Conference-Room-1  & 17.10 & 0.6247 & 0.6767 & 19.22$^\S$ & 0.6140$^\S$ & 0.6908$^\S$ \\
       Conference-Room-2 & 20.70 & 0.7267 & 0.5482 & 29.42 & 0.9444 & 0.1071  \\
       Kitchen-1   & 20.19 & 0.6957 & 0.5854 &  32.72 & 0.9454 & 0.1144  \\
       Kitchen-2  & 22.67 & 0.7538 & 0.4825 & 34.13 & 0.9634 & 0.0552  \\
       Robot-Workstation-1  & 20.37  & 0.7336 & 0.4149 &  22.90  & 0.8474 & 0.2020  \\
       Robot-Workstation-2  & 20.97 & 0.7420 & 0.4493 & 20.44 & 0.7779 & 0.3362 \\       
         \bottomrule
    \end{tabular}
    \begin{tablenotes}
      \small
      \item $^\S$ stops at epoch 4 due to convergence issues.
    \end{tablenotes}
    \end{threeparttable}
\end{table*}
\section{Related Works}
\label{sec:formatting}

\paragraph{One/Few-shot NeRF}
The field of one/few-shot Neural Radiance Fields (NeRF) has seen significant advancements in recent years. Several approaches have been proposed to tackle the challenges of synthesizing novel views and reconstructing 3D scenes with limited available data. 
Mip-NeRF 360~\cite{barron2022mipnerf360} addresses the task of Unbounded Anti-Aliased Neural Radiance Field synthesis, using a non-linear scene parameterization, online distillation, and a distortion-based regularizer to overcome the challenges presented by unbounded scenes.
EgoNeRF~\cite{choi2023balanced} employs spherical coordinates and leverages 360-degree panoramic videos as input to construct neural radiance fields.
In addition, there have been several multi-stage approaches~\cite{OmniPhotos,jang2022egocentric} that attempt to synthesize new view images by reconstructing an explicit mesh from egocentric omnidirectional videos. Zero-1-to-3~\cite{liu2023zero1to3} enables zero-shot novel view synthesis and 3D reconstruction using only a single image. 

In contrast, we assume a more casual input setting, where the viewpoint and scene composition may vary widely, data may be abundant, but from an egocentric viewpoint embedded in the scene, and video data is augmented by other multimodal data sources. 
\begin{table*}[t]
\begin{center}
\renewcommand\tabcolsep{0.5pt}
\begin{tabular}{>{\centering\arraybackslash}m{0.17\linewidth}>{\centering\arraybackslash}m{0.17\linewidth}>{\centering\arraybackslash}m{0.17\linewidth}>{\centering\arraybackslash}m{0.17\linewidth}>{\centering\arraybackslash}m{0.17\linewidth}>{\centering\arraybackslash}m{0.17\linewidth}}
\toprule
Step 0 & Step 50  & Step 100  & Step 150  & Step 200 \\
\midrule
\includegraphics[width=\linewidth]{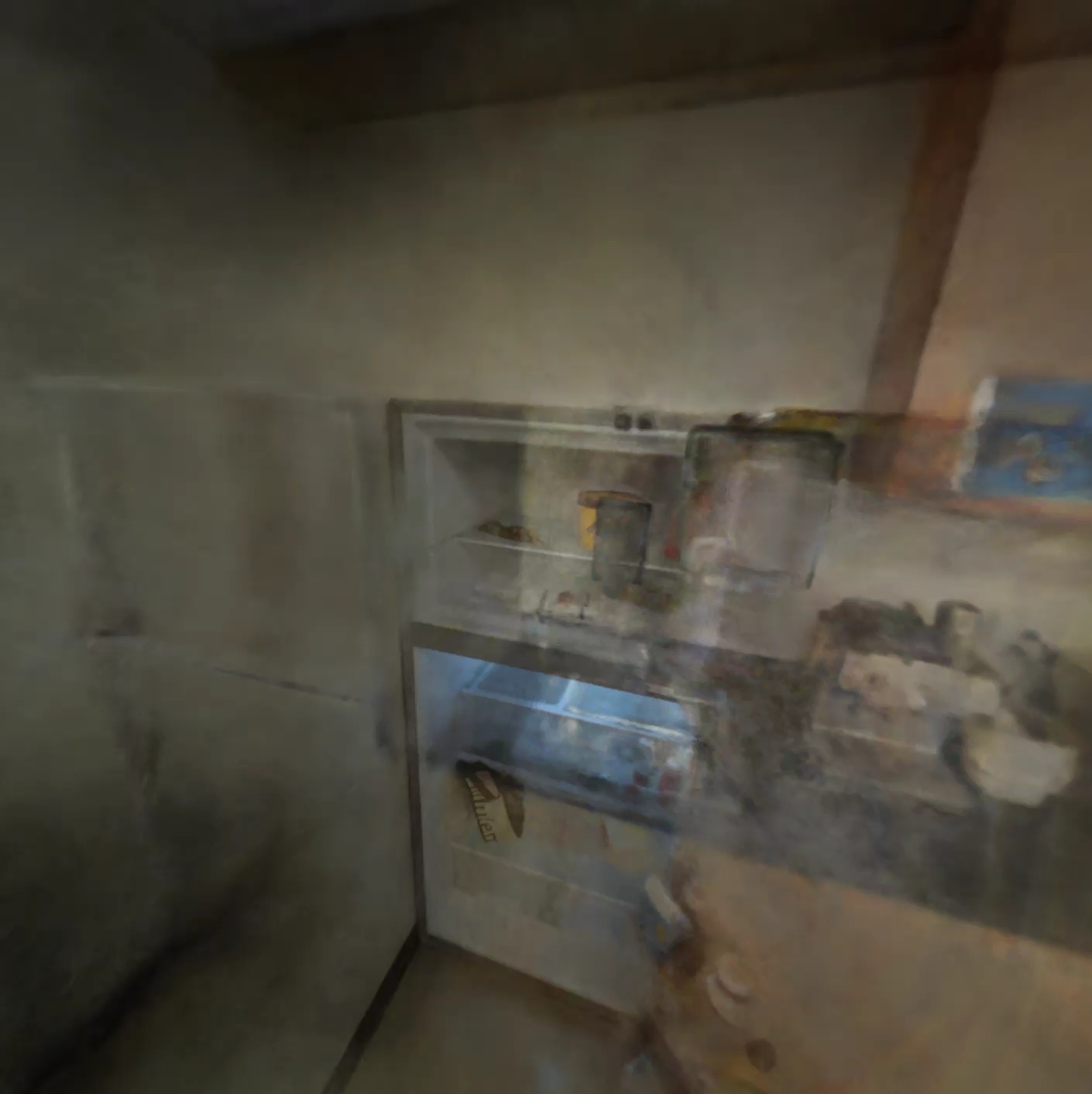} & \includegraphics[width=\linewidth]{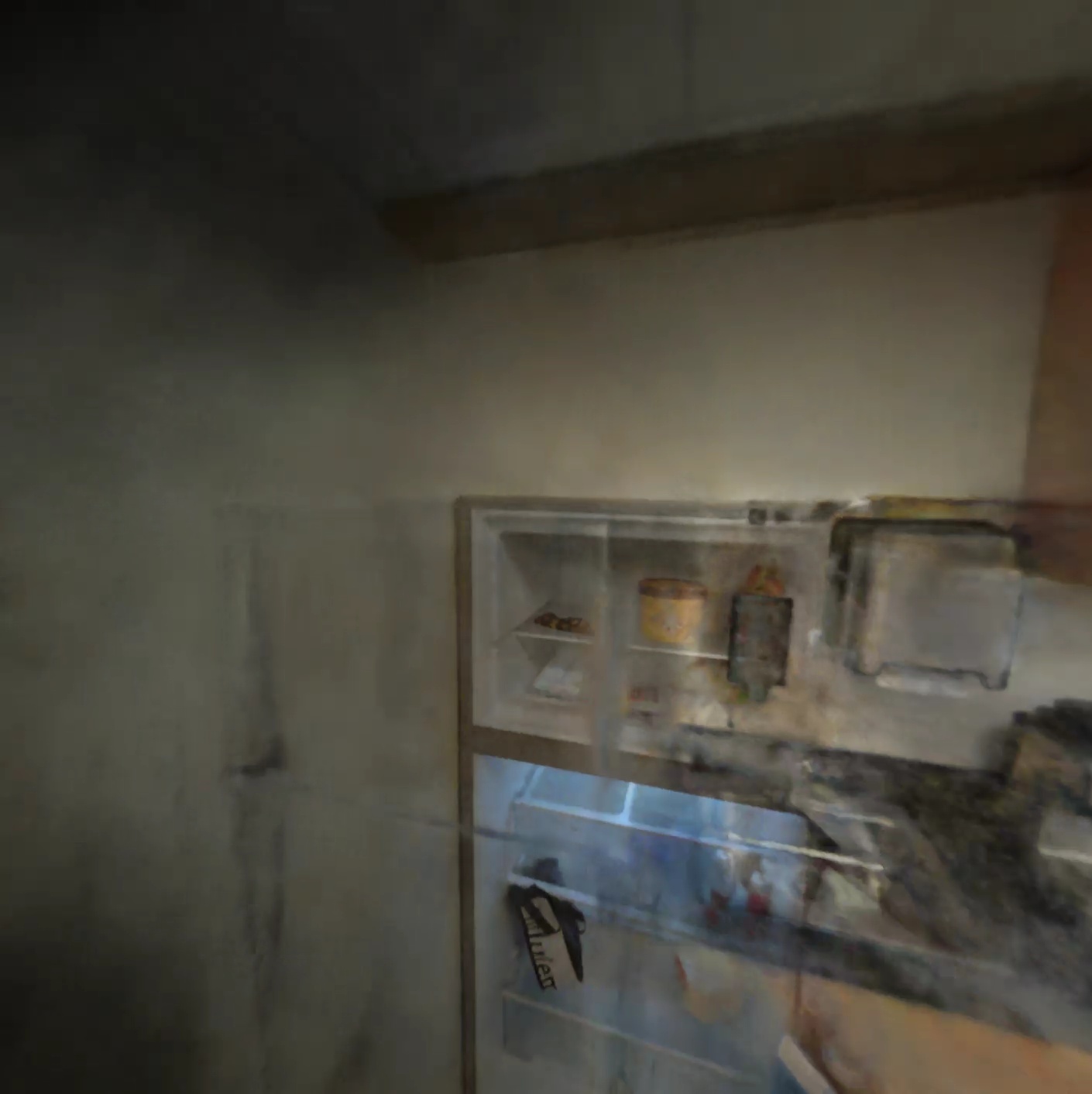}  & \includegraphics[width=\linewidth]{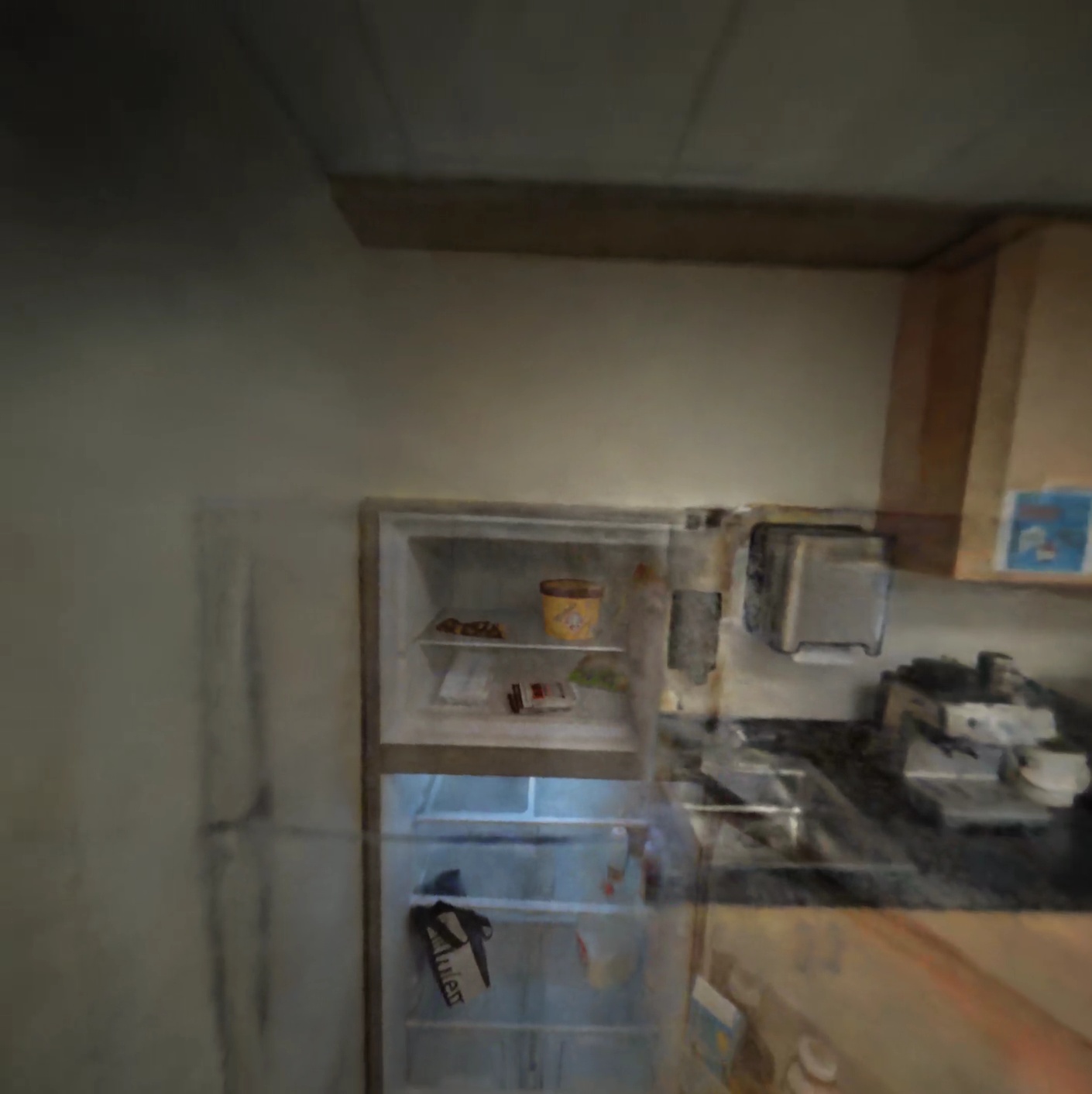} & \includegraphics[width=\linewidth]{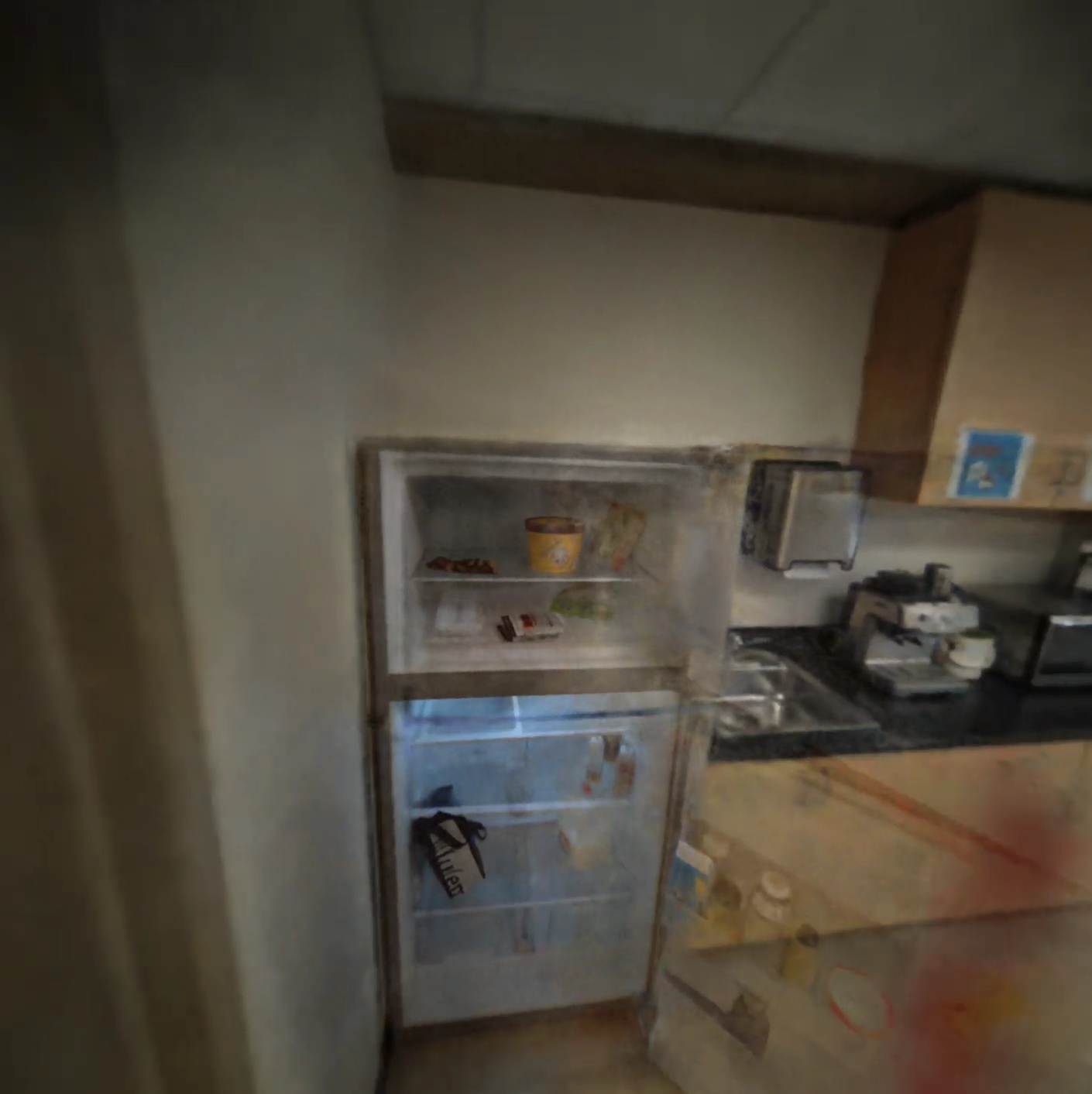} & \includegraphics[width=\linewidth]{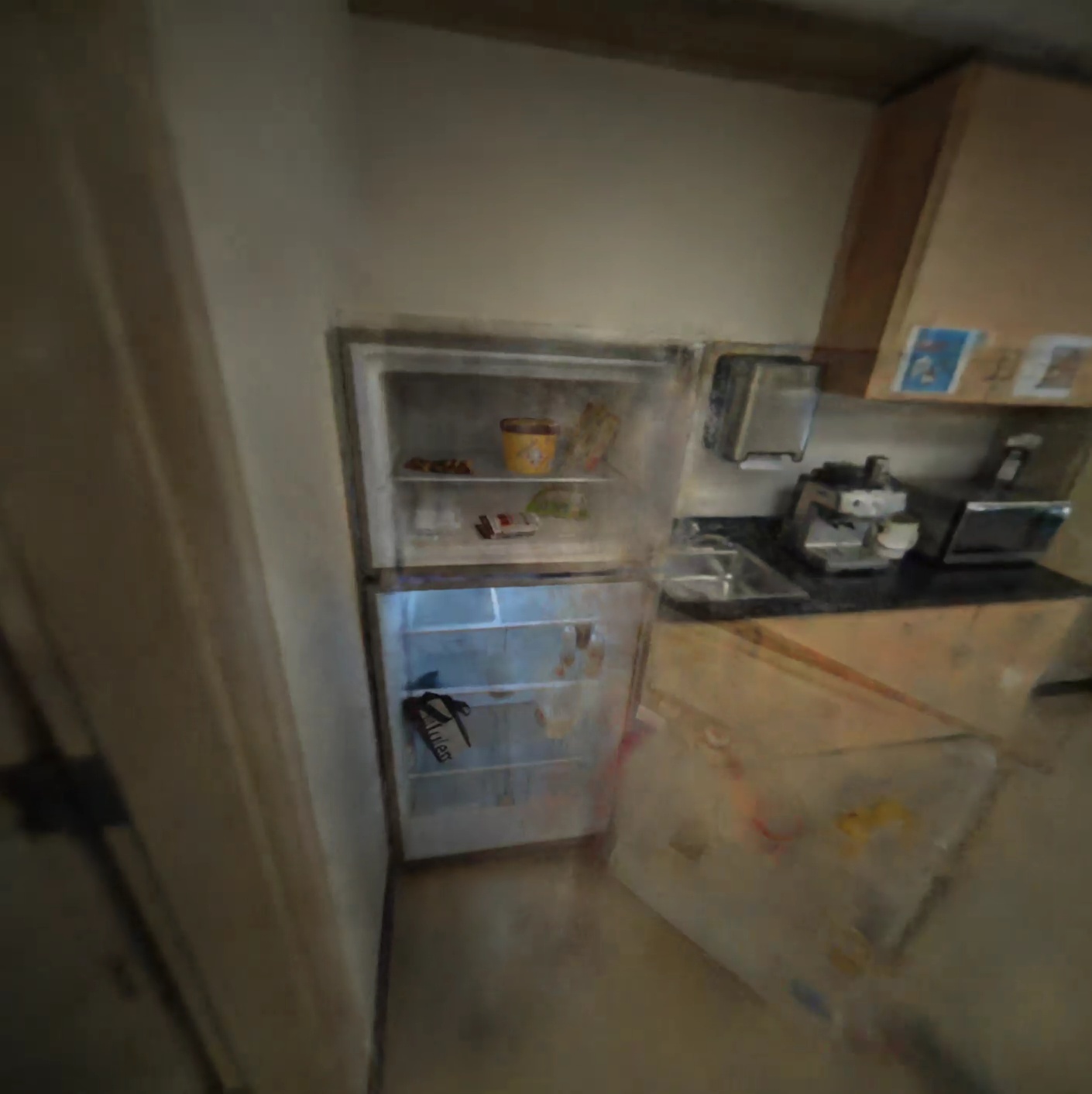} 
\\ 
Step 250 & Step 300  & Step 350  & Step 400  & Step 450 \\
\includegraphics[width=\linewidth]{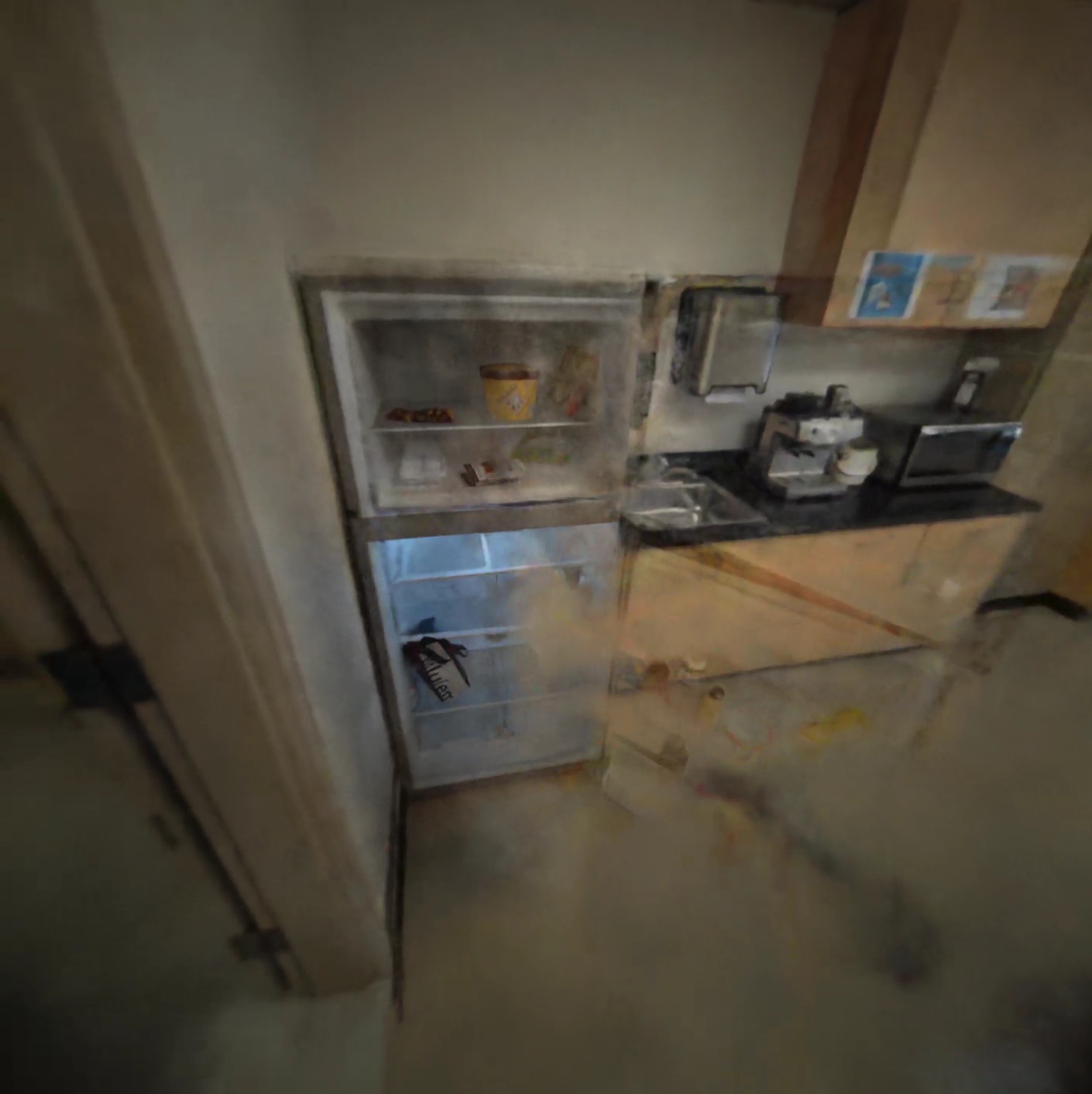} & \includegraphics[width=\linewidth]{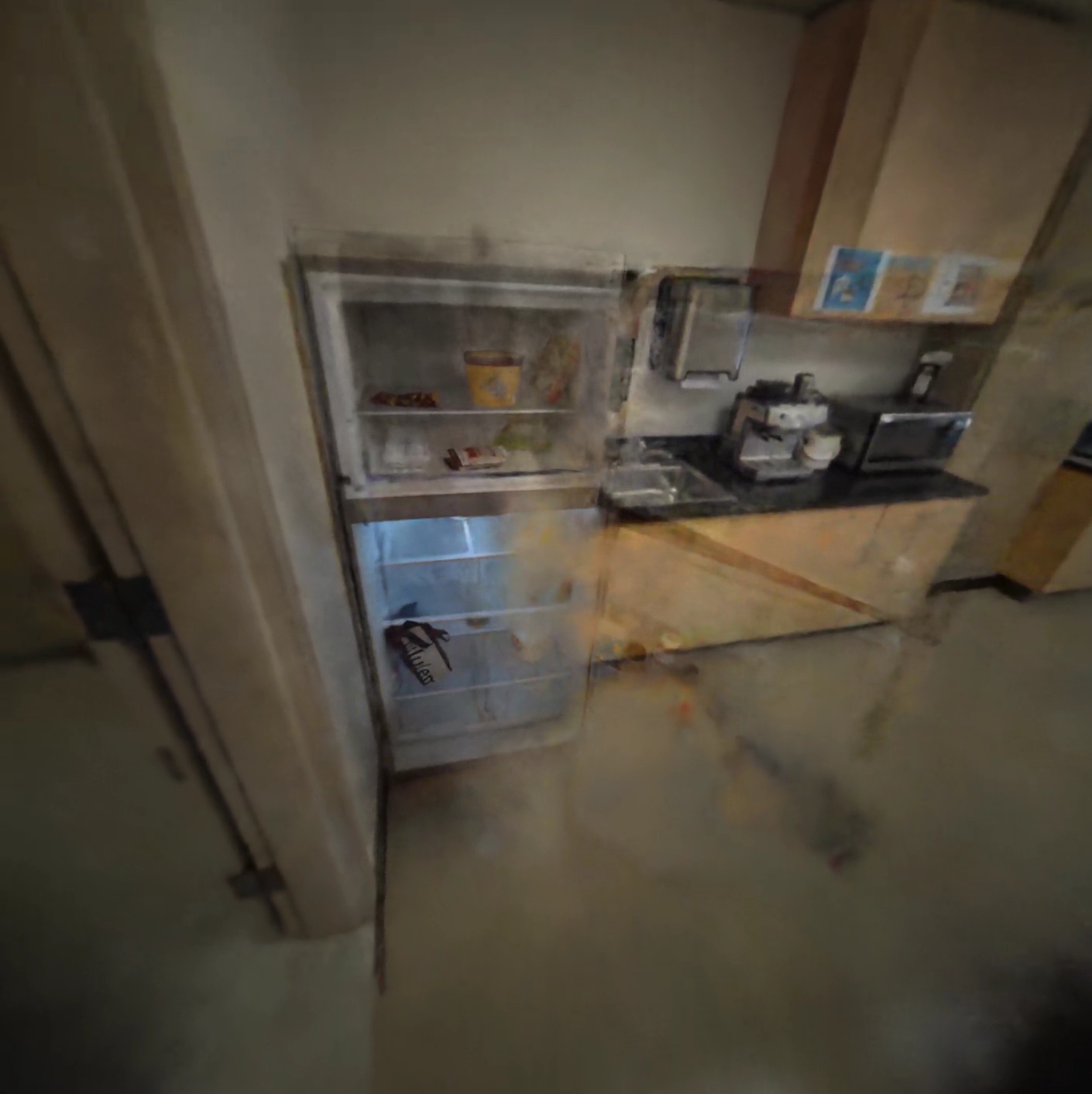}  & \includegraphics[width=\linewidth]{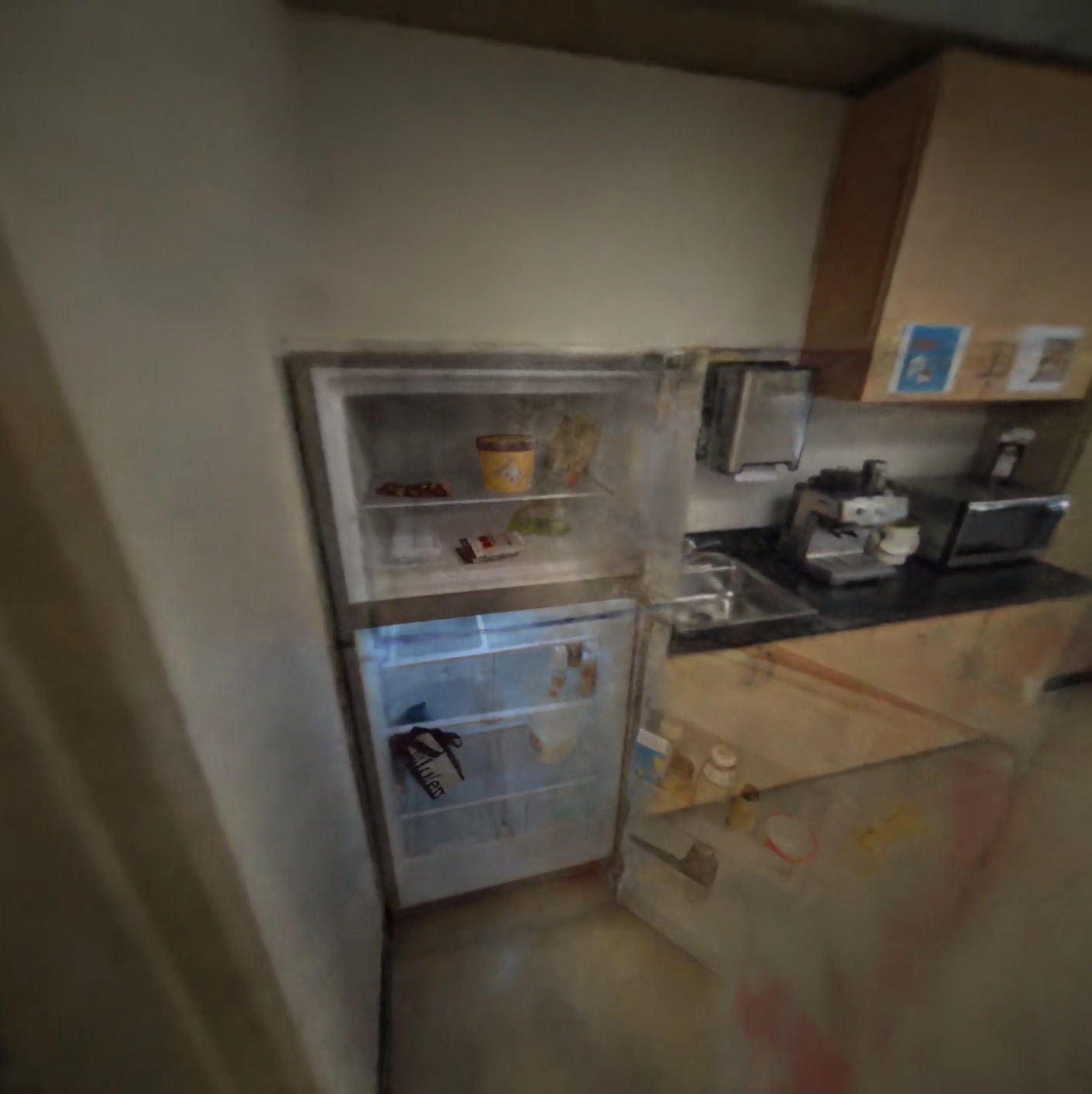} & \includegraphics[width=\linewidth]{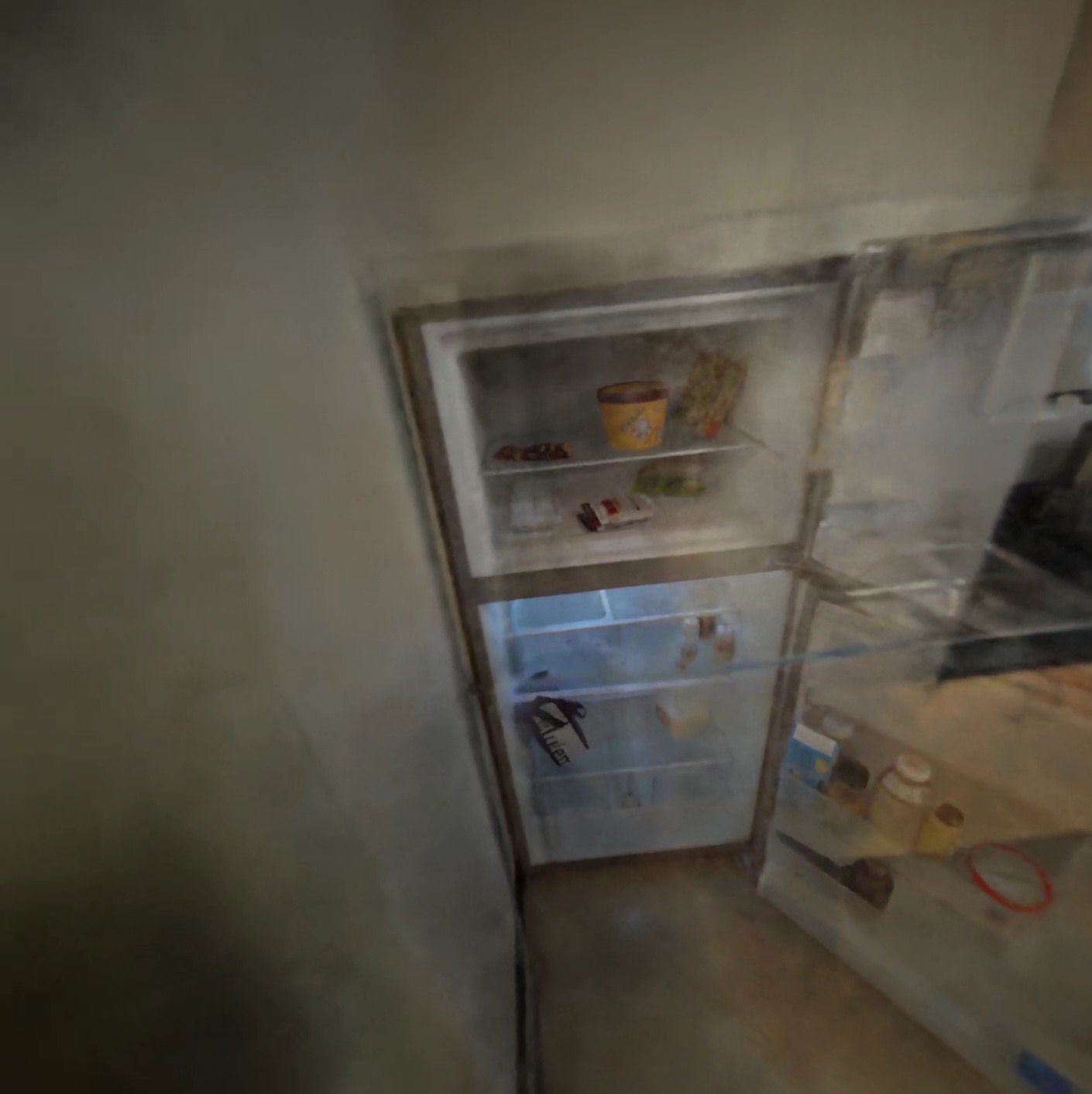} & \includegraphics[width=\linewidth]{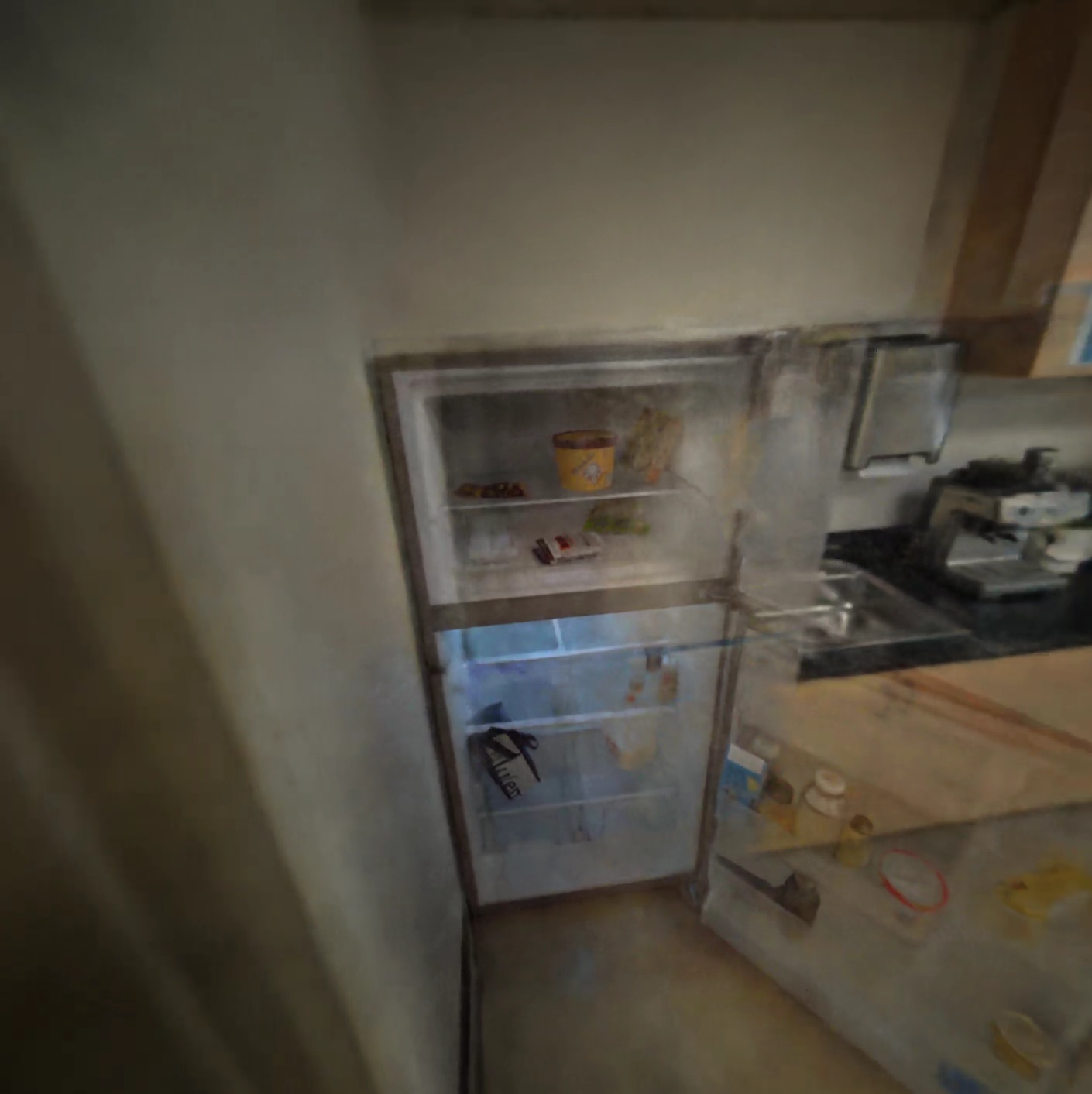} 
\\
\bottomrule
\end{tabular}
\captionof{figure}{Nerfacto Visualization Results on Kitchen 1 subset. We show the step numbers in the rendered video. Nerfacto results reveal some blurred regions, underscoring inherent limitations in its performance.}
\label{tab:nerfacto_kitchen_1}
\end{center}
\end{table*}
\paragraph{Dynamic NeRF} 
Recent studies have also focused on synthesizing novel views of dynamic scenes using a single camera. D-NeRF~\cite{pumarola2020d} has the capability to synthesize novel views of dynamic scenes with intricate non-rigid geometries at arbitrary time points. Nerfies~\cite{park2021nerfies} and HyperNeRF~\cite{park2021hypernerf} represent scenes using deformation fields that are conditioned on either time instant~\cite{pumarola2020d} or per-frame learned latent deformation
cod~\cite{park2021nerfies,park2021hypernerf,tretschk2021non}. NeuralDiff~\cite{tschernezki2021neuraldiff} tackles 3D object segmentation by employing a triple-stream neural renderer to separate the background, foreground, and actor. 
While these methods can handle lengthy videos, their primary effectiveness lies in object-centric scenes with limited object motion and controlled camera paths. Alternatively, some approaches model scenes as time-varying NeRFs~\cite{Gao-ICCV-DynNeRF,gao2022dynamic,li2021neural,wang2022fourier,xian2021space}.
NSFF~\cite{li2021neural} employs neural scene flow fields to capture complex 3D scene motion in real-world videos. However, it performs best on short, forward-facing videos lasting  1-2 seconds duration. 
DynIBaR~\cite{li2023dynibar}, focuses on synthesizing novel views from monocular videos depicting complex dynamic scenes. Our dataset is also well-suited for dynamic NeRF tasks.

\paragraph{Multimodal NeRF}
Multimodal Neural Radiance Field~\cite{zhan2023multimodal} is valuable for robot vision and scene understanding. Zhu et al.~\cite{zhu2023multimodal} introduce a method that aligns different modalities, incorporating point clouds and infrared image supervision. CLIP-NeRF~\cite{wang2022clip} proposes a unified framework that enables user-friendly manipulation of NeRF using either a short text prompt or an exemplar image. MMNeRF~\cite{zhang2023mmnerf} learns multimodal and multi-view features to guide neural radiance fields toward a generic model.
OMMO~\cite{lu2023large} serves as a multimodal benchmark for outdoor NeRF-based tasks, providing complex objects and scenes with calibrated images, point clouds, and prompt annotations. 
ObjectFolder~\cite{gao2022ObjectFolderV2} is a dataset designed for multisensory object-centric learning, incorporating vision, audio, and touch modalities. In addition to RGB cameras, our dataset includes a range of sensor types such as Fisheye cameras, IMU (Inertial Measurement Unit), audio, and more. This diverse collection of data enables the training of NeRF models using multiple modalities.

\paragraph{VR/AR}
NeRF holds great potential for creating immersive environments in Augmented and Virtual Reality (AR/VR) applications. NeRF is highly applicable with the ability to generate realistic and high-quality visual experiences in these domains.
Fov-NeRF~\cite{deng2022fov} is a technique that specifically targets Virtual Reality (VR) applications by introducing a gaze-contingent neural radiance field. This method enhances the responsiveness of neural synthesis within the VR environment, resulting in improved visual quality and realism in virtual experiences.
Instant-3D~\cite{li2023instant} is an algorithm-hardware co-design acceleration framework that enables instant on-device NeRF training. This framework facilitates instant 3D reconstruction for AR/VR applications, allowing for real-time and interactive experiences.
TLIO~\cite{liu2020tlio}, trained with pedestrian data from a headset, has the capability to produce statistically consistent measurements and uncertainty for IMU-only state estimation. This contributes to accurate tracking and positioning in AR/VR scenarios. EPIC Fields~\cite{EPICFields2023} enhances the EPIC-KITCHENS dataset by incorporating 3D camera information. Recently, HoloAssist~\cite{HoloAssist2023} is an egocentric human interaction dataset, where two people collaboratively complete physical manipulation tasks.
Our dataset, which includes a commodity omnidirectional camera with two fish-eye lenses, has the potential to enhance VR/AR applications by providing valuable data for training and improving NeRF-based techniques.

\begin{table*}[t]
\begin{center}
\renewcommand\tabcolsep{0.5pt}
\begin{tabular}{>{\centering\arraybackslash}m{0.12\linewidth}>{\centering\arraybackslash}m{0.17\linewidth}>{\centering\arraybackslash}m{0.17\linewidth}>{\centering\arraybackslash}m{0.17\linewidth}>{\centering\arraybackslash}m{0.17\linewidth}>{\centering\arraybackslash}m{0.17\linewidth}}
\toprule
 & Ground-truth & Composed Rendered  & Background  & Dynamic Foreground  & Actor \\
\midrule
Input View & \includegraphics[width=\linewidth]{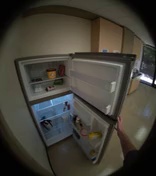} & \includegraphics[width=\linewidth]{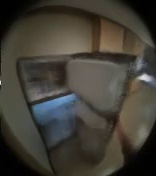}  & \includegraphics[width=\linewidth]{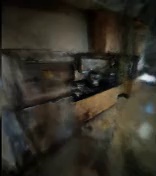} & \includegraphics[width=\linewidth]{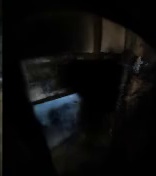} & \includegraphics[width=\linewidth]{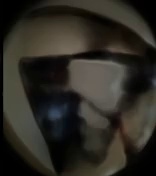} 
\\ 
Novel View & & \includegraphics[width=\linewidth]{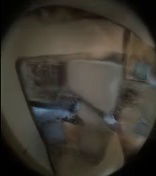}  & \includegraphics[width=\linewidth]{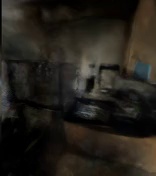} & \includegraphics[width=\linewidth]{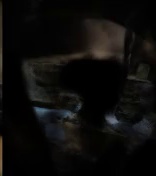} & \includegraphics[width=\linewidth]{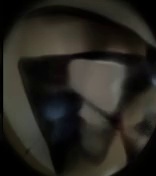} 
\\
\bottomrule
\end{tabular}
\captionof{figure}{NeuralDiff Visualization Results on Kitchen 1 subset. NeuralDiff can disentangle the background, dynamic foreground, and actors, all achieved in an unsupervised manner.}
\label{tab:neuraldiff_kitchen_1}
\end{center}
\end{table*}

\section{Method}
We evaluate two existing methods: Nerfacto~\cite{nerfstudio} which is adept at constructing neural radiance fields for static scenes using real-world data, and NeuralDiff~\cite{tschernezki2021neuraldiff}, a method explicitly tailored for dynamic Neural Radiance Field (NeRF) scenarios.
\subsection{Nerfacto}

The Nerfacto model is the default model used by nerfstudio~\cite{nerfstudio} for building neural radiance fields of static scenes from real data. This model is a  combination of various established methods known for their efficacy with real data. Key techniques integrated into the Nerfacto model include camera pose refinement, per-image appearance conditioning, proposal sampling, scene contraction, and hash encoding.

\paragraph{Ray Generation and Sampling}

The Nerfacto algorithm begins by optimizing camera views via an optimized SE(3) transformation~\cite{lin2021barf}. Utilizing these views, RayBundles are generated. To enhance both the efficiency and efficacy of the sampling process, a piece-wise sampler is employed. Initially, this sampler operates uniformly up to a designated distance from the camera. Subsequently, its sampling becomes progressively distributed, with each sample's step size increasing incrementally. These samples are introduced to a proposal network sampler, as conceptualized in the MipNeRF-360 approach~\cite{barron2022mip}. 
Nerfacto incorporates a compact fused MLP equipped with hash encoding~\cite{mueller2022instant} for representing the scene's density function, attributing to its computational efficiency without compromising accuracy. To further reduce the number of samples along rays, the proposal network sampler is designed to encompass multiple density fields. 

\paragraph{Scene Contraction and NeRF Field}

Many real-world scenes are unbounded, meaning they could extend indefinitely. 
Nerfacto applies scene contraction to transform this unbounded space into a fixed-size bounding box~\cite{barron2022mip}. Instead of the conventional $L^2$ norm contraction, Nerfacto adopts an $L^\infty$ norm contraction, resulting in a cubic domain rather than a spherical boundary. This cubic form is more conducive to the voxel-based hash encodings. Subsequently, these compacted spatial samples are compatible with the hash encoding framework provided by Instant-NGP, accessible through the tiny-cuda-nn~\cite{tiny-cuda-nn} Python interface. Additionally, Nerfacto integrates per-image appearance embeddings to mitigate variations in lighting and exposure encountered across different training cameras, referencing techniques used in~\cite{martinbrualla2020nerfw}. It also incorporates strategies from Ref-NeRF~\cite{verbin2022refnerf} to enhance the computation and prediction of surface normals.

\subsection{NeuralDiff}
NeuralDiff~\cite{tschernezki2021neuraldiff} is crafted for dynamic NeRF applications. It possesses the capability to autonomously disentangle the background, foreground, and actor within the NeRF representation.

NeuralDiff contains three sub-networks: Background density $\sigma_k^b\in \mathbb{R}_+$ and color $c_k^b\in \mathbb{R}^3$ can be obtained from the Background Network: $(\sigma_k^b, c_k^b) = \textrm{MLP}^b(g_tr_k, d_t)$.
The dynamic foreground is modelled by Foreground Network: $(\sigma_k^f, c_k^f, \beta_k^f) = \textrm{MLP}^f(g_tr_k, z_t^f)$  produces a ‘foreground’ occupancy $\sigma^f$ and color $c^f$ . Additionally, it predicts an uncertainty score $\beta^f_k$. A frame-specific code $z^f\in \mathbb{R}^D$ captures
the properties of the foreground that change over time. $z_t=B(t)\Gamma$ where $B(t)\in\mathbb{R}^P$ is a simple handcrafted basis and the motion $\Gamma\in\mathbb{R}^{P\times D}$ are coefficients such that $P \ll T$. Specifically, $B(t)=[1, t, \sin 2\pi t, \cos 2\pi t, \sin 4\pi t, \cos 4\pi t, \cdots]$ is a deterministic harmonic coding of time. Foreground objects are manipulated by the actor/observer, whose movements are sporadic, while the actor's body undergoes continuous motion. To model this dynamic actor, Actor Network $(\sigma_k^a, c_k^a, \beta_k^a) = \textrm{MLP}^f(r_k, z_t^a)$ is used. The key difference is that the 3D point $r_k$ is expressed relative to the camera (v.s. $g_tr_k$ which is expressed relative to the world). $g_t\in SE(3)$ is the moving camera motion, where $SE(3)$ is the group of Euclidean transformations. $d_t$ is the unit-norm viewing direction. 

\subsection{Dataset and Benchmark}
Our dataset possesses three distinct characteristics:
\begin{itemize}
    \item It comprises egocentric and dynamic scenes.
    \item The data is derived from real-world scenes. 
    \item It incorporates multiple modalities.
\end{itemize}

The data modalities in our \datasetName are captured by RGB cameras, ET camera, Microphone, Barometer, GPS, Wi-Fi, Bluetooth, SLAM/Mono Scene camera left, SLAM/Mono Scene camera right, IMU (1kHz), IMU (800Hz), and Magnetometer. The statistics for each subset and for the different modalities can be found in Tab.~\ref{tab:dataset_stat1} and Tab.~\ref{tab:dataset_stat2}.

\subsubsection{Data Collection}
Our dataset includes different scenarios. We collected multimodal sensory data in these scenarios, including RGB videos, ET camera, Microphone, Barometer, GPS, Wi-Fi, Bluetooth, SLAM, IMU (1kHz), IMU (800Hz), and Magnetometer, as shown in Figure~\ref{fig:dataset_intro}. Each participant wears Aria Glasses to perform specific tasks within each scenario. These tasks may include activities like navigating, utilizing tools, and interacting with common household objects. In certain scenarios, GPS information is unavailable due to the absence of GPS signals within indoor environments. Similarly, the collection of Bluetooth data information is contingent upon the presence of a Bluetooth device near the scene. In the absence of a nearby Bluetooth device, capturing Bluetooth data becomes unfeasible.

In our data preprocessing pipeline, we employ Aria Data Tools\footnote{\url{https://facebookresearch.github.io/Aria_data_tools/}} to extract data from MPS files\footnote{\url{https://facebookresearch.github.io/projectaria_tools/docs/data_utilities/core_code_snippets/mps}}, which is later utilized for visualization purposes. For pose estimation based on RGB video sequences, we employ COLMAP~\cite{schoenberger2016sfm}. Our dataset does not necessitate additional annotations.

\section{Experiments}
In this section, we first introduce the training details. We then evaluate Nerfacto and NeuralDiff on our dataset.
Our analysis reveals that the proposed \datasetName is challenging, and there is much room for current NeRF methods to improve in the context of egocentric view synthesis.
\subsection{Baselines and Implementation Details}
We run two baselines on our collected dataset:
\begin{itemize}
    \item Nerfacto: We employed the default training settings of nerfstudio, conducting training for $30,000$ iterations with an initial learning rate of $10^{-8}$ during the pre-warmup phase, and a final learning rate of $0.0001$.
    \item NeuralDiff: The NeuralDiff model is trained for 10 epochs with a learning rate of $0.0005$, utilizing 64 ray samples.
\end{itemize}

\subsection{Quantitative Results}

Qualitative results are presented in Table~\ref{tab:quan_results} for Nerfacto and NeuralDiff. In terms of PSNR, SSIM, and LPIPS metrics, NeuralDiff generally surpasses Nerfacto across various scenarios. However, it is important to note that Nerfacto produces a de-distorted image, while NeuralDiff generates a fisheye image closely resembling the ground truth. The latter exhibits curvature characteristics.

\subsection{Qualitative Results}

Qualitative results are presented in Figure~\ref{tab:nerfacto_kitchen_1} for Nerfacto and Figure~\ref{tab:neuraldiff_kitchen_1} for NeuralDiff. Notably, the visualization of Nerfacto reveals some blurred regions, underscoring inherent limitations in its performance. In contrast, the visualization of NeuralDiff demonstrates its remarkable ability to disentangle elements within the scene, disentangling the background, dynamic foreground, and actors, all in an unsupervised manner. Both methods exhibit potential for enhancement in the application of dynamic NeRF.

\section{Discussion}
An advantage of Aria Glasses, in contrast to the HoloLens, is their lightweight and highly portable design, making them seamlessly adaptable to people's everyday lives.
It is worth noting that a limitation of the Aria Glasses RGB sensor is its relatively lower image resolution compared to current smartphone RGB cameras. 

Recent advances in foundation models have brought new paradigm shifts and breakthroughs in different research areas~\cite{bommasani2021opportunities,qiu2023large}. Foundation models emerge with generalist intelligence that can solve a wide range of tasks after being trained with a large quantity of data. With data at its core, foundation model research is embracing a new trend towards multimodality~\cite{zhang2023meta,qiu2023visionfm}. \datasetName, along with other large-scale egocentric datasets such as  Ego4D~\cite{grauman2022ego4d} and EPIC-KITCHENS~\cite{Damen2021PAMI}, holds great promise in bolstering the development of multimodal foundation models for egocentric view synthesis. Once trained, Aria-NeRF can also be used for many downstream perception and planning tasks, such as NeRF-based object detection~\cite{10175562}, semantic segmentation~\cite{Zhi:etal:ICCV2021}, and so on.

\section{Conclusion}
In this work, we tackled the problem of egocentric view synthesis. To facilitate research in this field, we introduced \datasetName, a multimodal egocentric dataset captured using Aria Glasses. We benchmarked two baseline models, Nerfacto and NeuralDiff, on this novel dataset. While these two models can generate reasonable view synthesis, the experimental results revealed much potential for improvement given the challenging nature of the dataset. The dataset's rich diversity of modalities and real-world context lay a solid groundwork for advancing our understanding of human behavior and bolstering more immersive and intelligent experiences in the realms of VR and AR.

{\small
\bibliographystyle{ieee_fullname}
\bibliography{egbib}
}

\end{document}